\documentclass[11pt,a4paper]{article}
\usepackage[hyperref]{emnlp-2020-template/emnlp2020}
\usepackage{times}
\usepackage{latexsym}

\usepackage{microtype}

\usepackage{setspace}

\aclfinalcopy 
\def\aclpaperid{2936} 

\usepackage{amsmath}

\aclfinalcopy 
\def\aclpaperid{2936} 

\usepackage{xargs}                      

\usepackage{multirow}
\usepackage{arydshln}
\usepackage{graphicx}
\usepackage{float}
\usepackage{subcaption}
\graphicspath{ {./img/} }

\usepackage{verbatim}
\usepackage{bbm}

\title{Finding the Optimal Vocabulary Size for Neural Machine Translation}

\author{ Thamme Gowda and Jonathan May \\
  Information Sciences Institute\\
   University of Southern California\\
 {\tt \{tg, jonmay\}@isi.edu}\\
 }

\date{}

\begin{document}
\maketitle


\begin{abstract}

We cast neural machine translation (NMT) as a classification task in an autoregressive setting and analyze the limitations of both classification and autoregression components.
Classifiers are known to perform better with balanced class distributions during training.
Since the Zipfian nature of languages causes imbalanced classes, we explore its effect on NMT. 
We analyze the effect of various vocabulary sizes on NMT performance on multiple languages with many data sizes, and reveal an explanation for \textit{why} certain vocabulary sizes are better than others.\footnote{Tools, configurations, system outputs, and analyses are at \href{https://github.com/thammegowda/005-nmt-imbalance}{https://github.com/thammegowda/005-nmt-imbalance}}

\end{abstract}

\section{Introduction}

Natural language processing (NLP) tasks such as sentiment analysis \cite{maas-etal-2011-learning, Zhang-etal-15-cnn-sentiment} and spam detection are modeled as classification tasks, where instances are independently labeled.
Tasks such as part-of-speech tagging \cite{CoNLL2017-shared-UD} and named entity recognition \cite{CoNLL-2003-NER} are examples of structured classification tasks, where instance classification is decomposed into a sequence of per-token contextualized labels.
We can similarly cast neural machine translation (NMT), an example of a natural language generation (NLG) task, as a form of structured classification, where an instance label (a translation) is generated as a sequence of contextualized labels, here by an autoregressor (see Section \ref{sec:classifier-nlg}).

Since the parameters of modern machine learning (ML) classification models are estimated from training data, whatever biases exist in the training data will affect model performance.
Among those biases, \textit{class imbalance} is a topic of our interest. 
Class imbalance is said to exist when one or more classes are not of approximately equal frequency in data.
The effect of class imbalance has been extensively studied in several domains where classifiers are used (see Section \ref{sec:rel-class-imb}).
With neural networks, the imbalanced learning problem is mostly targeted to computer vision tasks; NLP tasks are under-explored \cite{Johnson2019SurveyImbalance}. 
 
Word types in natural language models resemble a Zipfian distribution, i.e. in any natural language corpus, we observe that a type's rank is roughly inversely proportional to its frequency. Thus, a few types are extremely frequent, while most of the rest lie on the long tail of infrequency. 
Zipfian distributions cause two problems in classifier-based NLG systems:
\begin{enumerate}
    \itemsep0em 
    \item \textbf{Unseen Vocabulary:} 
    Any hidden data set may contain types not seen in the finite set used for training. A sequence drawn from a Zipfian distribution is likely to have a large number of rare types, and these are likely to have not been seen in training.
    \item \textbf{Imbalanced Classes:} There are a few extremely frequent types and many infrequent types, causing an extreme imbalance.  
    Such an imbalance, in other domains where classifiers are used, has been known to cause undesired biases and severe performance degradation \cite{Johnson2019SurveyImbalance}. 
\end{enumerate} 

The use of \textit{subwords}, that is, decomposition of word types into pieces, such as the widely used Byte Pair Encoding (BPE) \cite{sennrich-etal-2016-bpe} addresses the open-ended vocabulary problem by ultimately allowing a word to be represented as a sequence of characters if necessary.
BPE has a single hyperparameter named \textit{merge operations} that governs the vocabulary size. 
The effect of this hyperparameter is not well understood. 
In practice, it is either chosen arbitrarily or via trial-and-error \cite{DBLP:journals/corr/abs-1810-08641}.

Regarding the problem of imbalanced classes, \newcite{steedman-2008-last} states that ``the machine learning techniques that we rely on are actually very bad at inducing systems for which the crucial information is in rare events.''
However, to the best of our knowledge, this problem has not yet been directly addressed in the NLG setting.

In this work, we attempt to find answers to these questions: \textit{`What value of BPE vocabulary size is best for NMT?'}, and more crucially an explanation for \textit{`Why that value?'}.
As we will see, the answers and explanations for those are an immediate consequence of a broader question, namely \textit{`What is the impact of Zipfian imbalance on classifier-based NLG?'}

The contributions of this paper are as follows:
We offer a simplified view of NMT architectures by re-envisioning them as two high-level components: a \textit{classifier} and an \textit{autoregressor} (Section~\ref{sec:classifier-nlg}).
We describe some of the desired settings for the classifier (Section~\ref{sec:classifier-balance}) and autoregressor (Section~\ref{sec:ar-short-seq}) components.
In Section~\ref{sec:bpe}, we describe how vocabulary size choice relates to the desired settings for the two components. 
Our experimental setup is described in Section~\ref{sec:exp-setup}, followed by an analysis of results in Section~\ref{sec:nmt_analysis} that offers an explanation with evidence for \textit{why} some vocabulary sizes are better than others.  
Section~\ref{sec:class-bias} uncovers the impact of class imbalance, particularly frequency based discrimination on classes.\footnote{In this work, `type' and `class' are used interchangeably.}
Section~\ref{sec:related-work} provides an overview of  related work, and in Section~\ref{sec:conclusion} we recommend a heuristic for choosing the BPE hyperparameter.

\section{Classifier based NLG}
\label{sec:classifier-nlg}

\begin{figure}[ht]
    \centering
    \includegraphics[width=0.8\linewidth]{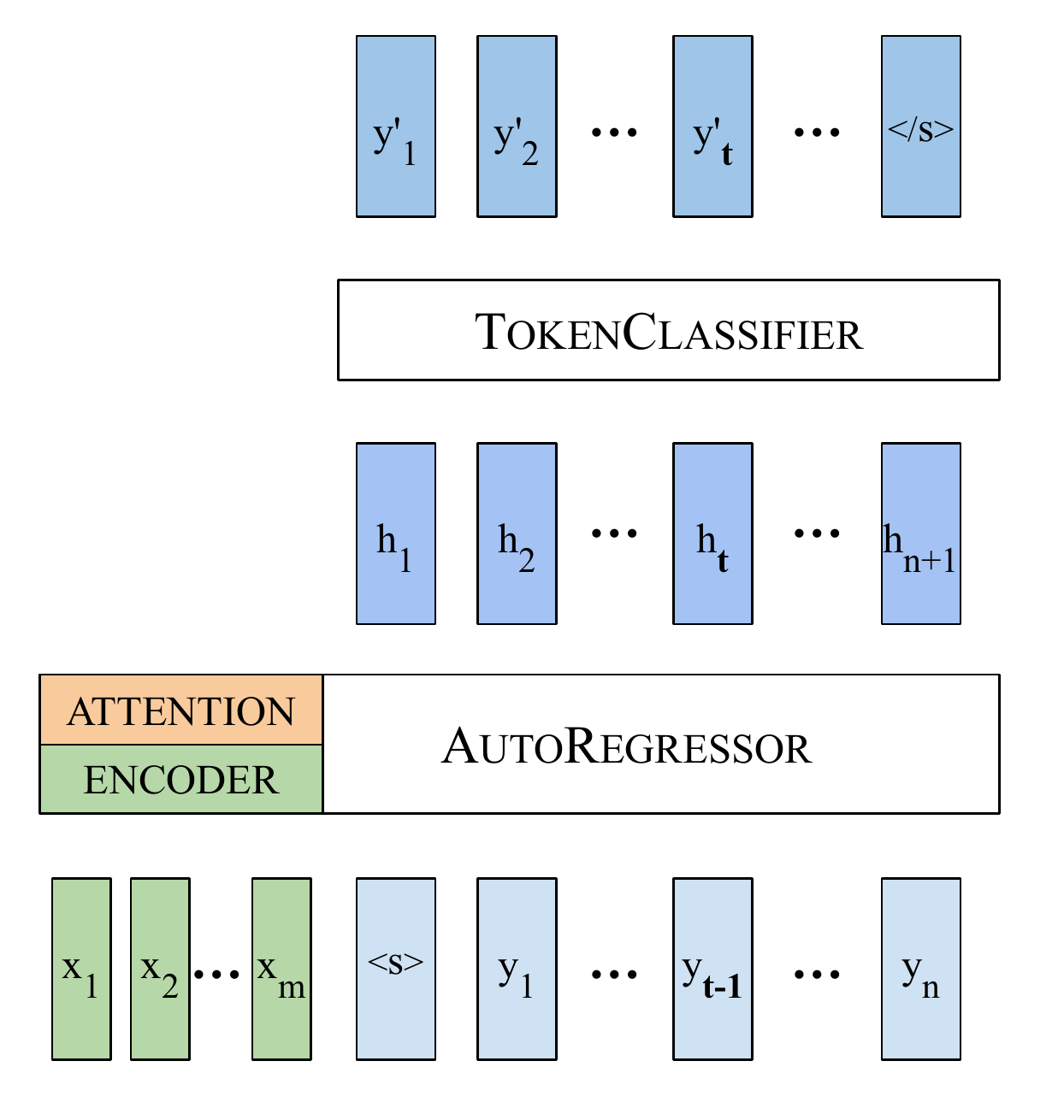}
    \caption{The NMT model re-envisioned as a token classifier with an autoregressive feature extractor.}
    \label{fig:nmt-architecture}
\end{figure}

Machine translation is commonly defined as the task of transforming sequences from the form $x = x_1 x_2 x_3 ... x_m$ to $y = y_1 y_2 y_3 ... y_n$, where $x$ is in source language $X$ and $y$ is in target language $Y$. 
There are many variations of NMT architectures (Section \ref{sec:rel-nmt-arch}), however, all share the common objective of maximizing ${ \prod_{t=1}^{n} P(y_t | y_{<t}, x_{1:m})}$ for pairs $(x_{1:m}, y_{1:n})$ sampled from a parallel dataset. 
NMT architectures are commonly viewed as encoder-decoder networks.
We instead re-envision the NMT architecture as two higher level components: an autoregressor ($R$) and a token classifier ($C$), as shown in Figure~\ref{fig:nmt-architecture}.

Autoregressor $R$, \cite{box2015time} being the most complex component of the NMT model, has many implementations based on various neural network architectures: recurrent neural networks (RNN) such as long short-term memory (LSTM) and gated recurrent unit (GRU), convolutional neural networks (CNN), and Transformer (Section \ref{sec:rel-nmt-arch}). 
At time step $t$, $R$ transforms the input context $y_{<t}, x_{1:m}$ into hidden state vector $h_t = R(y_{<t}, x_{1:m})$.

Classifier $C$ is the same across all architectures.
It maps $h_t$ to a distribution $P(y_j | h_t) \forall y_j \in V_Y$, where $V_Y$ is the vocabulary of $Y$. 
In machine learning, input to classifiers such as $C$ is generally described as features that are either hand-engineered or automatically extracted.
In our high-level view of NMT architectures, $R$ is a neural network that serves as an automatic feature extractor for $C$.

\subsection{Balanced Classes for Token Classifier}
\label{sec:classifier-balance}

Untreated, class imbalance leads to bias based on class frequencies.
Specifically, classification learning algorithms focus on frequent classes while paying relatively less importance to infrequent classes.
Frequency-based bias leads to poor recall of infrequent classes \cite{Johnson2019SurveyImbalance}. 

When a model is used in a \textit{domain mismatch} scenario, i.e. where test and training set distributions do not match, model performance generally degrades.
It is not surprising that frequency-biased classifiers show particular degradation in domain mismatch scenarios, as  types that were infrequent in the training distribution and were ignored by the learning algorithm may appear with high frequency in the new domain.
\newcite{koehn2017sixchallenges} showed empirical evidence of poor generalization of NMT to out-of-domain datasets.

In other classification tasks, where each instance is classified independently, methods such as up-sampling infrequent classes and down-sampling frequent classes are used.
In NMT, since classification is done within the context of sequences, it is possible to accomplish the objective of balancing by altering sequence lengths.
This can be done by choosing the level of subword segmentation \cite{sennrich-etal-2016-bpe}.

\textbf{Quantification of Zipfian Imbalance:}
We use two statistics to quantify the imbalance of a training distribution:

The first statistic relies on a measure of \textbf{Divergence} ($D$) from a balanced (uniform) distribution. 
We use a simplified version of Earth Mover Distance, in which the total cost for moving a probability mass between any two classes  is the sum of the total mass moved.
Since any mass moved \textit{out of} one class is moved \textit{into} another, we divide the total per-class mass moves in half to avoid double counting.  
Therefore, the imbalance measure $D$ on $K$ class distributions where $p_i$ is the observed probability of class $i$ in the training data is computed as:
$$D = \frac{1}{2} \sum_{i=1}^{K}| p_i - \frac{1}{K}|; \quad 0 \le D \le 1 $$

A lower value of $D$ is the desired setting for $C$, since the lower value results from a balanced class distribution. 
When classes are balanced, they have approximately equal frequencies; $C$ is thus less likely to make errors due to class bias.

The second statistic is \textbf{Frequency at 95th\% Class Rank (\textbf{$F_{95\%}$})}, defined as the least frequency in the $95^{th}$ percentile of most frequent classes.
More generally, $F_{\large{P\%}}$ is a simple way of quantifying the minimum number of training examples for at least the$P$th percentile of classes.
The bottom $(1-P)$ percentile of classes are overlooked to avoid the noise that is inherent in the real-world natural-language datasets.

A higher value for $F_{95\%}$ is the desired setting for $C$, as a higher value indicates the presence of many training examples per class, and ML methods are known to perform better when there are many examples for each class.

\subsection{Shorter Sequences for Autoregressor}
\label{sec:ar-short-seq}

Every autoregressive model is an approximation; some may be better than others, but no model is perfect. 
The total error accumulated grows in proportion to the length of the sequence.
These accumulated errors alter the prediction of subsequent tokens in the sequence.
Even though beam search attempts to mitigate this, it does not completely resolve it.  
These challenges with respect to long sentences and beam size are examined by \newcite{koehn2017sixchallenges}.

We summarize sequence lengths using \textbf{Mean Sequence Length}, $\mu$, 
computed trivially as the arithmetic mean of the lengths of \textit{target} language sequences after encoding them:
$\mu = \frac{1}{N} \sum_{i=1}^N |y^{(i)}|$
where $y^{(i)}$ is the $i$th sequence in the training corpus of $N$ sequences.
Since shorter sequences have relatively fewer places where an imperfectly approximated autoregressor model can make errors, a smaller $\mu$ is a desired setting for $R$.

\subsection{Choosing the Vocabulary Size Systematically}
\label{sec:bpe}

BPE~\cite{sennrich-etal-2016-bpe} is a greedy iterative algorithm often used to segment a vocabulary into useful \textit{subwords}. 
The algorithm starts with characters as its initial vocabulary.
In each iteration, it greedily selects the most frequent type bigram in the training corpus, and replaces the sequence with a newly created compound type.
Once the subword vocabulary is learned, it can be applied to a corpus by greedily segmenting words with the longest available subword type. These operations have an effect on $D$, $F_{95\%}$, and $\mu$.

\textbf{Effect of BPE on $\mu$}:
BPE expands rare words into two or more subwords, lengthening a sequence (and raising $\mu$) relative to simple white-space segmentation.
BPE merges frequent-character sequences into one subword piece, shortening a sequence (and lowering $\mu$) relative to character segmentation.
Hence, the sequence length of BPE segmentation lies in between the sequence lengths obtained by white-space and character-only segmentation methods \cite{morishita-etal-2018-improving}. 

\textbf{Effect of BPE on $F_{95\%}$ and $D$}:
Whether BPE is viewed as a merging of frequent subwords into a relatively less frequent compound, or a splitting of rare words into relatively frequent subwords, BPE alters the class distribution by moving the probability mass of classes.
Hence, by altering the class distribution, BPE also alters both $F_{95\%}$ and $D$. The BPE hyperparameter controls the amount of probability mass moved between subwords and compounds.

\begin{figure}[ht]
  \centering
    \includegraphics[width=\linewidth]{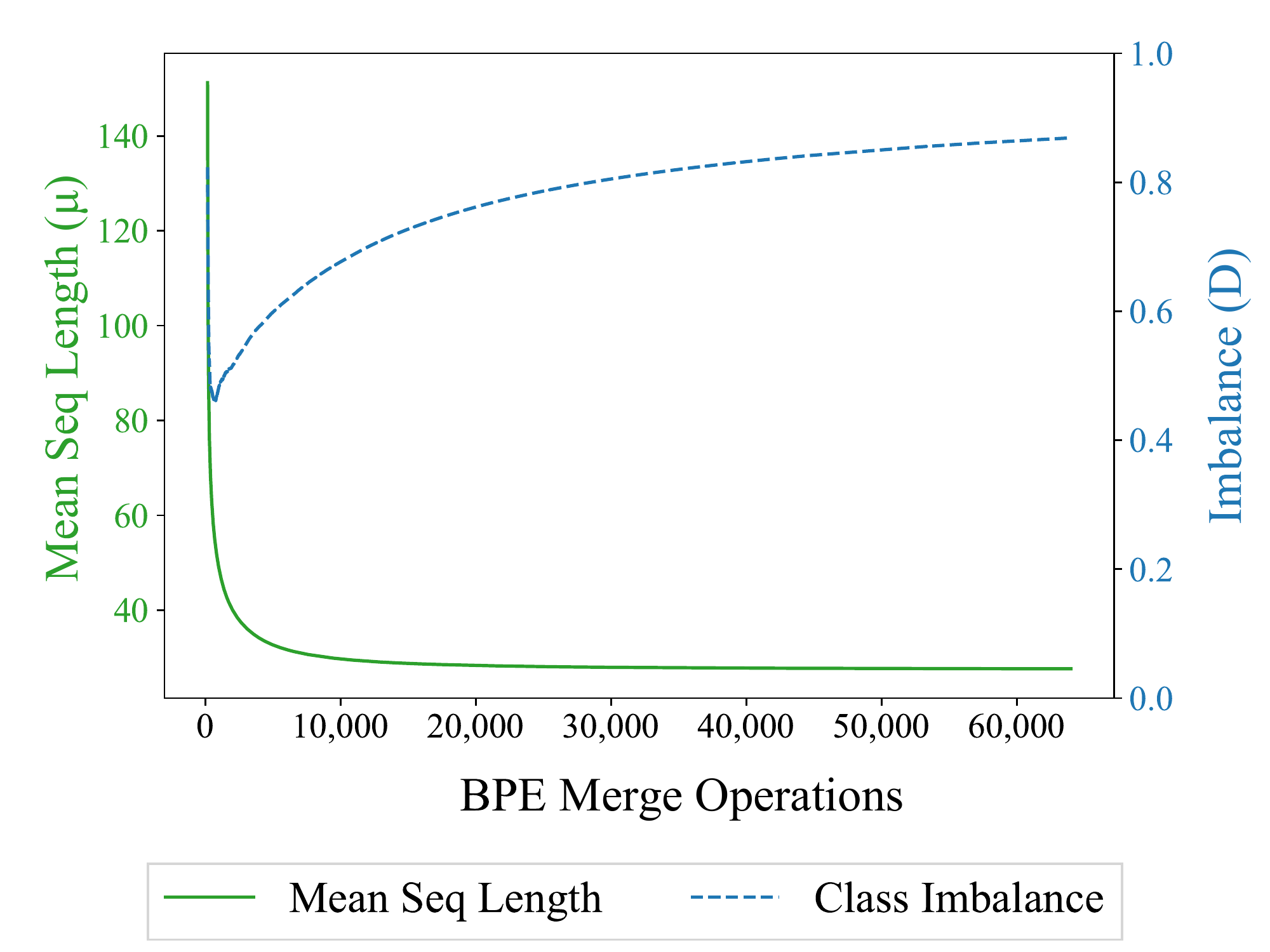}
    \caption{Effect of BPE merge operations on mean sequence length ($\mu$) and class imbalance ($D$).
    } 
    \label{fig:BPE-imbalance}
\end{figure}

Figure~\ref{fig:BPE-imbalance} shows the relation between number of BPE merges (i.e. the BPE hyperparameter), and both $D$ and $\mu$.
When few BPE merge operations are performed, we observe the lowest value of $D$, which is a desired setting for $C$, but at the same point $\mu$ is large and undesired for $R$ (Section~\ref{sec:classifier-nlg}).
When a large number of BPE merges are performed, the effect is reversed, i.e. we observe that $D$ is large and unfavorable to $C$ while $\mu$ is small and favorable to $R$. 
In the following sections we describe our experiments and analysis to locate the optimal number of BPE merges that achieves the right trade-off for both $C$ and $R$. 

\section{Experimental Setup}
\label{sec:exp-setup}

Our NMT experiments use the base Transformer model \cite{vaswani2017attention} on four different target languages at various training data sizes, described in the following subsections. 

\subsection{Datasets}
We use the following four language pairs for our analysis: English$\rightarrow$German, German$\rightarrow$English, English$\rightarrow$Hindi, and English$\rightarrow$Lithuanian. 
To analyze the impact of different training data sizes, we randomly sub-select smaller training corpora for English$\leftrightarrow$German and English$\rightarrow$Hindi languages. 
Statistics regarding the corpora used for validation, testing, and training are in Table~\ref{tab:datasets}.
The datasets for English$\leftrightarrow$German, and English$\rightarrow$Lithuanian are retrieved from the News Translation task of WMT2019~\cite{wmt19proceedings}.\footnote{\href{http://www.statmt.org/wmt19/translation-task.html}{http://www.statmt.org/wmt19/translation-task.html}}
For English$\rightarrow$Hindi, we use the IIT Bombay Hindi-English parallel corpus v1.5~\cite{iitb-hien}.
English, German, and Lithuanian sentences are tokenized using \textsc{SacreMoses}.\footnote{\href{https://github.com/alvations/sacremoses}{https://github.com/alvations/sacremoses}} 
Hindi sentences are tokenized using \textsc{IndicNlpLibrary}.\footnote{\href{https://github.com/anoopkunchukuttan/indic_nlp_library}{https://github.com/anoopkunchukuttan/indic\_nlp\_library}}

The training datasets are trivially cleaned: we exclude sentences with length in excess of five times the length of their parallel counterparts. 
Since the vocabulary is a crucial part of this analysis, we exclude all sentence pairs containing URLs. 

\begin{table*}[ht]
    \centering
\begin{tabular}{ l  l  r  r  r  l  l }
  Languages & Training & Sentences & EN Toks & XX Toks & Validation & Test \\ \hline
\multirow{4}{1.5cm}{DE$\rightarrow$EN\\EN$\rightarrow$DE} 
    & \multirow{4}{3.5cm}{Europarl v10 \\ WMT13CommonCrawl \\ NewsCommentary v14}
         & 30K  &  0.8M & 0.8M & \multirow{4}{*}{ {\small NewsTest18} }
            & \multirow{4}{*}{ \small{NewsTest19}} \\ \cdashline{3-5}
    &    & 0.5M  & 12.9M & 12.2M &  &  \\ \cdashline{3-5}
    &    & 1M    & 25.7M & 24.3M &  &  \\ \cdashline{3-5} 
    &    & 4.5M  & 116M & 109.8M &  &  \\ \hdashline

\multirow{2}{*}{EN$\rightarrow$HI } 
     & \multirow{2}{*}{ IITB Training } 
         & 0.5M & 8M & 8.6M  & \multirow{2}{*}{ \small{IITB Dev} } 
     & \multirow{2}{*}{ \small{IITB Test} }  \\\cdashline{3-5} 
     &   & 1.3M & 21M & 22.5M   &   &  \\ \hdashline 
EN$\rightarrow$LT & Europarl v10 & 0.6M & 17M & 13.4M  & \small{NewsDev19} & \small{NewsTest19} \\
\end{tabular} 
    \caption{Training, validation, and testing datsets, along with sentence and token counts in training sets. We generally refer to dataset's sentence size in this work.}
    \label{tab:datasets}
\end{table*}

\subsection{Hyperparameters}
Our model is a 6 layer Transformer encoder-decoder that has 8 attention heads, 512 hidden vector units, and a feed forward intermediate size of 2048, with GELU activation. 
We use label smoothing at 0.1, and a dropout rate of 0.1.
We use the Adam optimizer \cite{kingma2015adam} with a controlled learning rate that warms up for 16K steps followed by the decay rate recommended for training Transformer models~\cite{popel2018tfm-train-tips}. 
To improve performance at different data sizes we set the mini-batch size to 6K tokens for the 30K-sentence datasets, 12K tokens for 0.5M-sentence datasets, and 24K for the remaining larger datasets~\cite{popel2018tfm-train-tips}. 
All models are trained until no improvement in validation loss is observed, with a patience of 10 validations, each done at 1,000 update steps apart. 
Our model is implemented using PyTorch and run on NVIDIA P100 and V100 GPUs.
To reduce padding tokens per batch, mini-batches are made of sentences having similar lengths \cite{vaswani2017attention}.
We trim longer sequences to a maximum of 512 tokens after BPE.
To decode, we average the last 10 checkpoints, and use a beam size of 4 with length penalty of 0.6, similar to \newcite{vaswani2017attention}.

Since the vocabulary size hyperparameter is the focus of this analysis, we use a range of vocabulary sizes that include character vocabulary and BPE operations that yield vocabulary sizes between 500 and 64K types.
A common practice, as seen in \newcite{vaswani2017attention}'s setup, is to jointly learn BPE for both source and target languages, which facilitates three-way weight sharing between the encoder's input, the decoder's input, and the output (i.e. classifier's class) embeddings \cite{press-wolf-2017-using}.
However, to facilitate fine-grained analysis of vocabulary sizes and their effect on class imbalance, our models separately learn source and target vocabularies; weight sharing between the encoder's and decoder's embeddings is thus not possible.
For the target language, however, we share weights between the decoder's input and the classifier's class embeddings.

\section{Results and Analysis}
\label{sec:nmt_analysis}
\begin{figure}[ht]
\begin{subfigure}{\linewidth}
    \centering
    \includegraphics[width=0.98\linewidth]{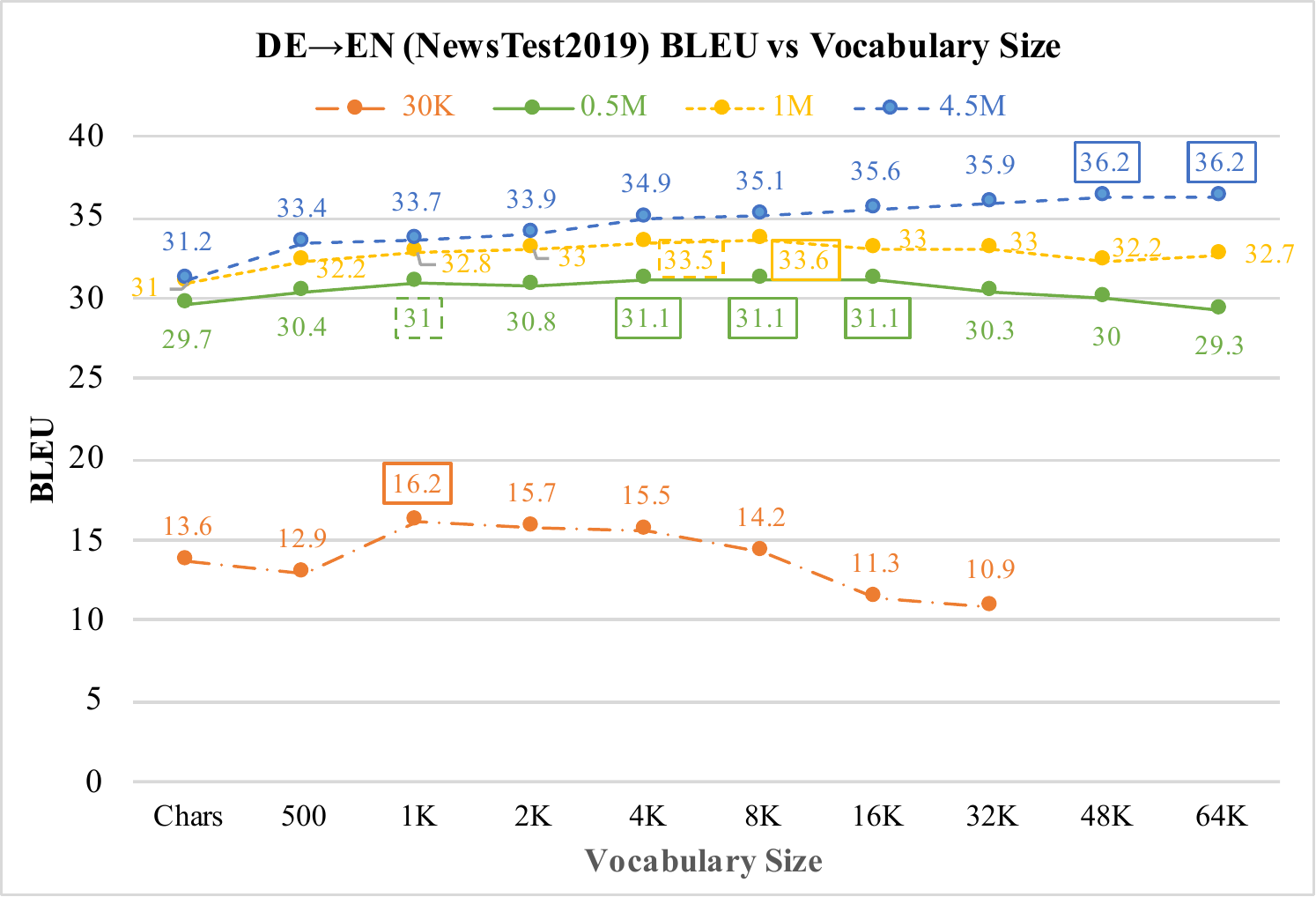}
    \caption{DE$\rightarrow$EN BLEU on NewsTest2019}
    \label{fig:bleu-deen}
\end{subfigure}
\begin{subfigure}{\linewidth}
    \centering
    \includegraphics[width=0.98\linewidth]{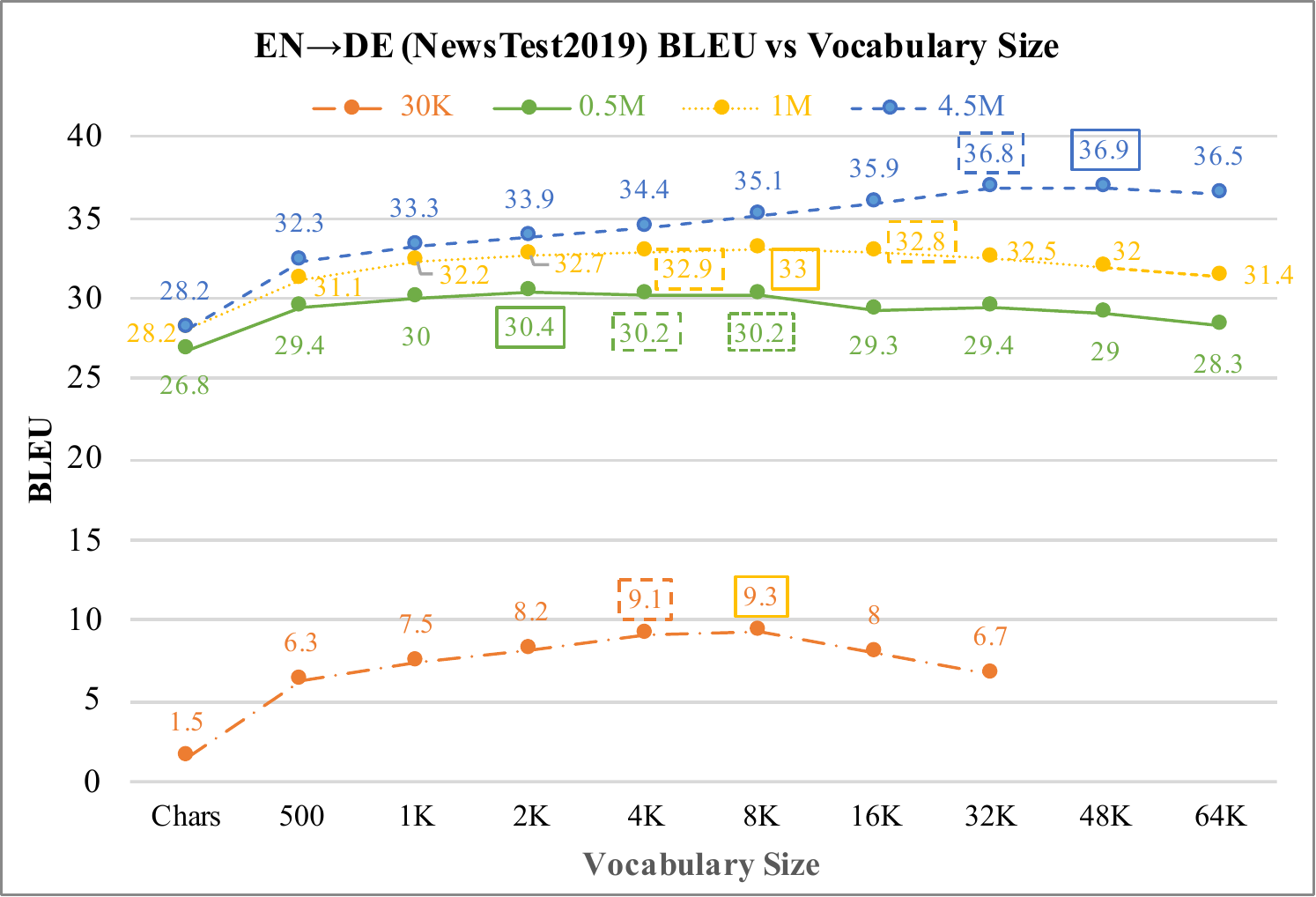}
    \caption{EN$\rightarrow$DE BLEU on NewsTest2019}
    \label{fig:bleu-ende}
\end{subfigure}
\caption{EN$\leftrightarrow$DE NewsTest2019 BLEU as a function of vocabulary size at various training set sizes. 
Only the large dataset with 4.5M sentences has its best performance at a large vocabulary; all others peak at an 8K or smaller vocabulary size.}
\label{fig:bleu-ende-deen}
\end{figure}

\begin{figure}[ht]
    \centering    
    \includegraphics[width=0.98\linewidth]{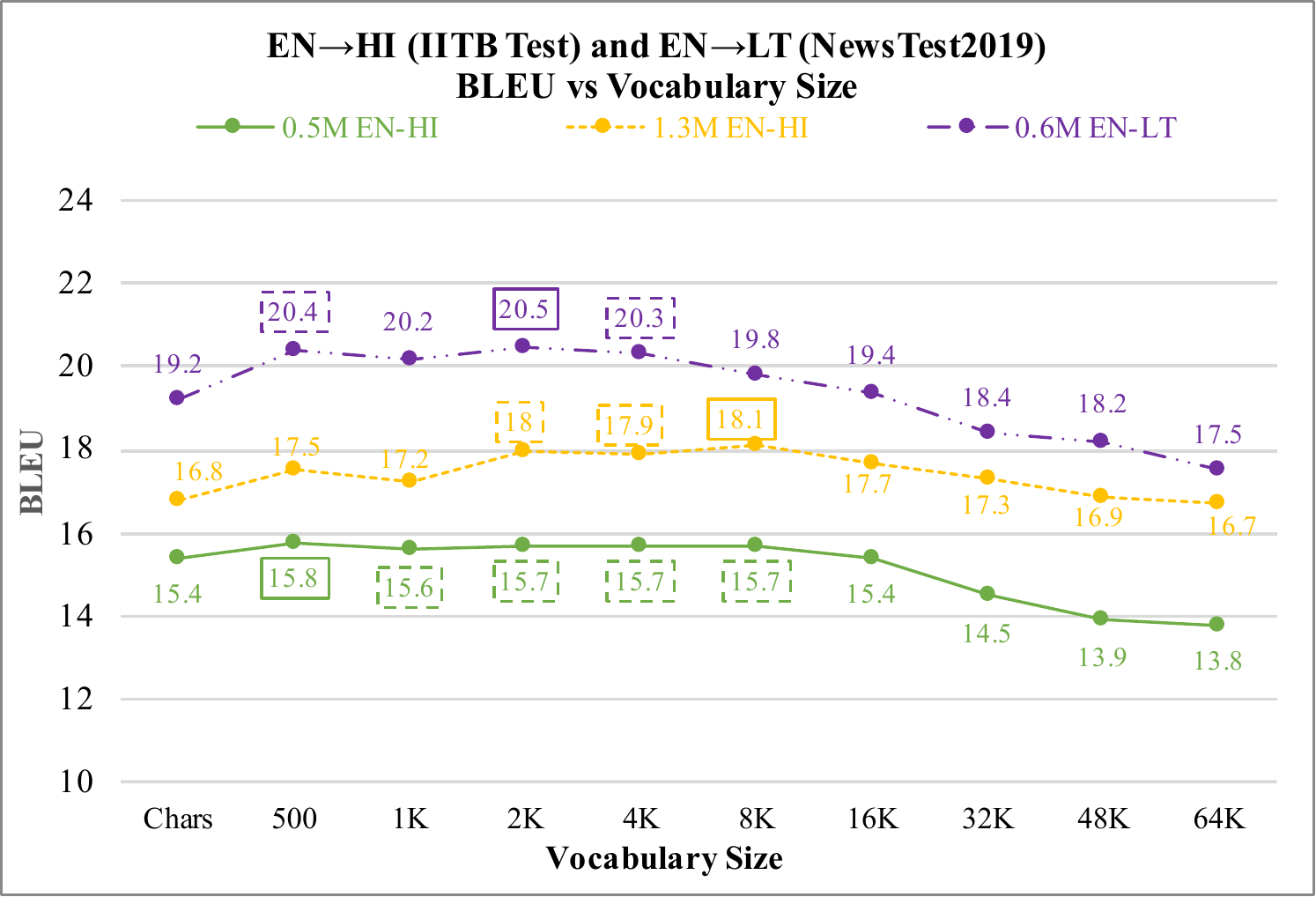}
    \caption{BLEU on EN$\rightarrow$HI IITB Test and EN$\rightarrow$LT NewsTest2019 as a function of vocabulary size.
    These language pairs observed the best BLEU scores in the range of 500 to 8K vocabulary size.}
    \label{fig:bleu-enhilt}
\end{figure}

BLEU scores for DE$\rightarrow$EN and EN$\rightarrow$DE experiments are reported in Figures~\ref{fig:bleu-deen} and \ref{fig:bleu-ende} respectively.
Results from EN$\rightarrow$HI, and EN$\rightarrow$LT are combined in Figure~\ref{fig:bleu-enhilt}.
All the reported BLEU scores are obtained using \textsc{SacreBLEU} \cite{post-2018-sacreBLEU}.\footnote{\texttt{BLEU+case.mixed+numrefs.1+smooth.exp+ tok.13a+version.1.4.6}}

We make the following observations: smaller vocabulary such as characters have not produced the best BLEU for any of our language pairs or dataset sizes. 
A vocabulary of 32K or larger is unlikely to produce optimal results unless the data set is large e.g. the 4.5M DE$\leftrightarrow$EN sets.
The BLEU curves as a function of vocabulary sizes have a shape resembling a hill. 
The position of the peak of the hill seems to shift towards a larger vocabulary when the datasets are large. 
However, there is a lot of variance in the position of the peak: one extreme is at 500 types on 0.5M EN$\rightarrow$HI, and the other extreme is at 64K types in 4.5M DE$\rightarrow$EN. 
    
\begin{figure*}[ht]
\begin{subfigure}{\textwidth}
  \centering
  \includegraphics[width=0.6\linewidth,trim={1.4cm 0 0.2cm 16.45cm},clip]{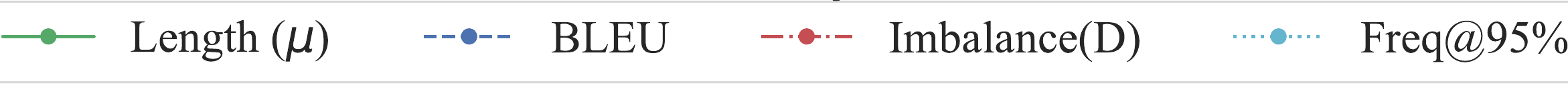}
\end{subfigure}

\begin{subfigure}{.33\textwidth}
  \centering
  \includegraphics[width=0.99\linewidth,trim={2.4cm 1.32cm 4.1cm 0},clip]{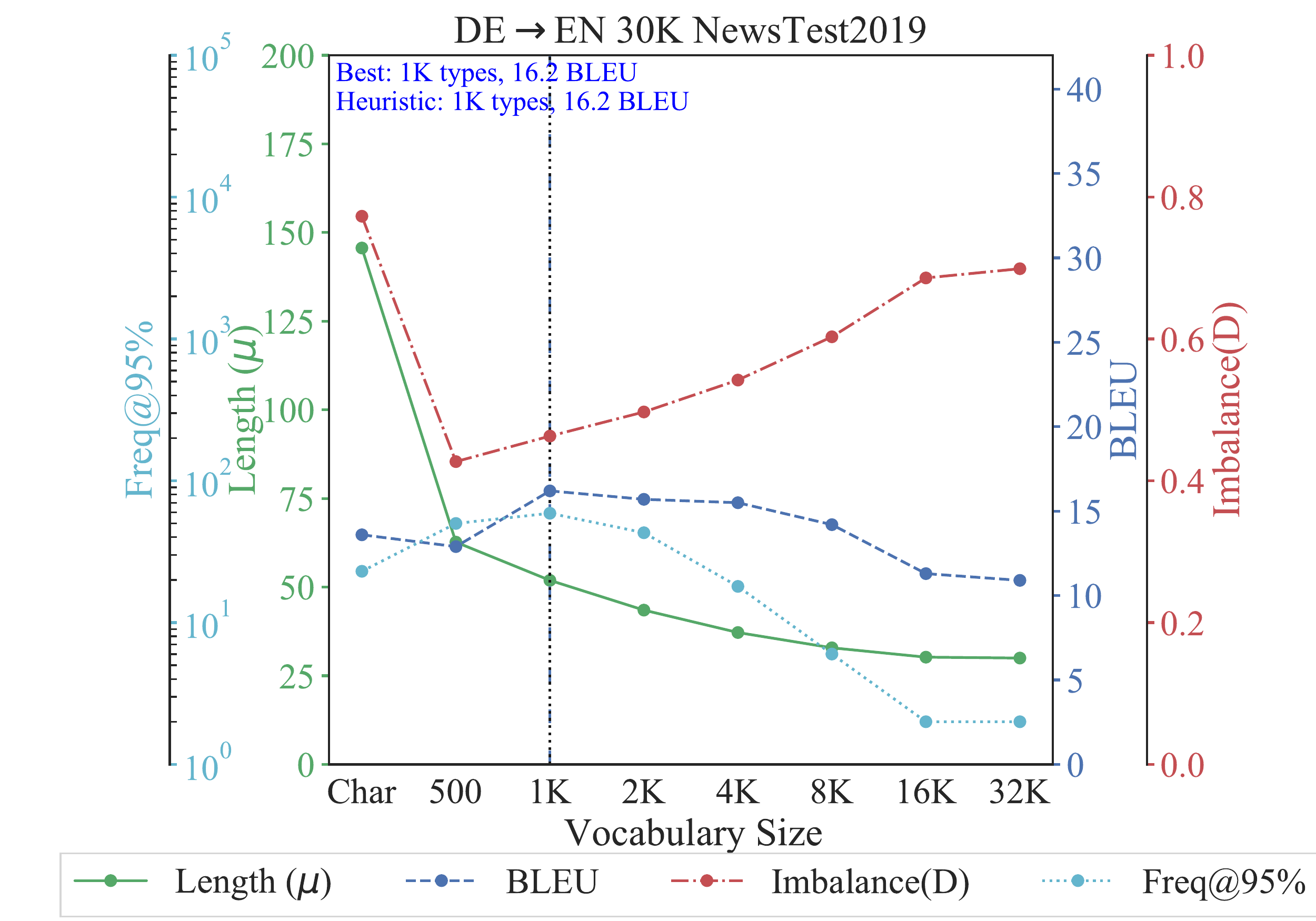}
\end{subfigure}
\begin{subfigure}{.32\textwidth}
  \centering
  \includegraphics[width=0.89\linewidth,trim={5.1cm 1.32cm 4.1cm 0},clip]{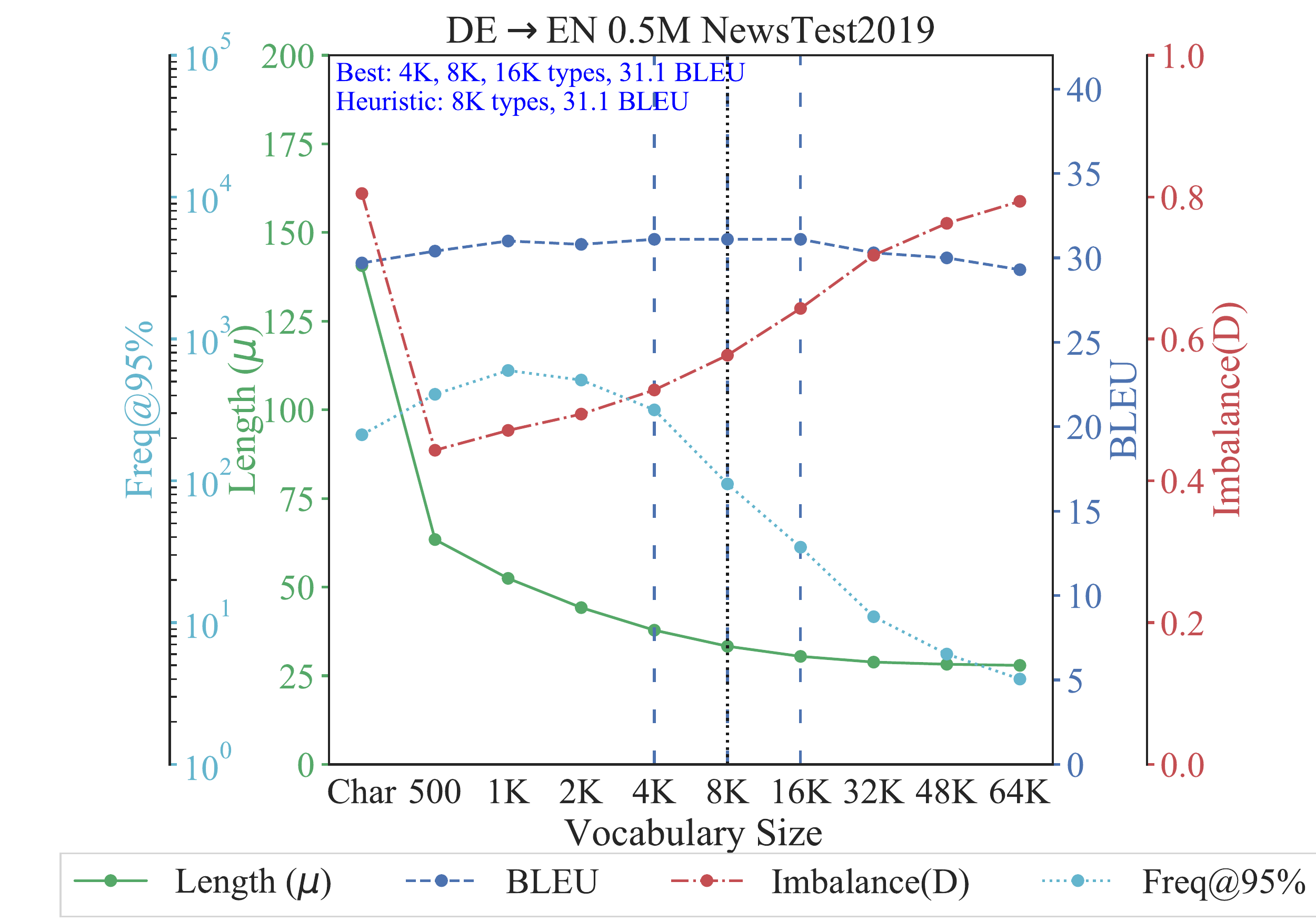}
\end{subfigure}
\begin{subfigure}{.33\textwidth}
  \centering
  \includegraphics[width=0.99\linewidth,trim={5.1cm 1.32cm 1.4cm 0},clip]{4axv-test-deen-4.5m.pdf}
\end{subfigure}

\begin{subfigure}{.33\textwidth}
  \centering
  \includegraphics[width=0.99\linewidth,trim={2.4cm 1.32cm 4.1cm 0},clip]{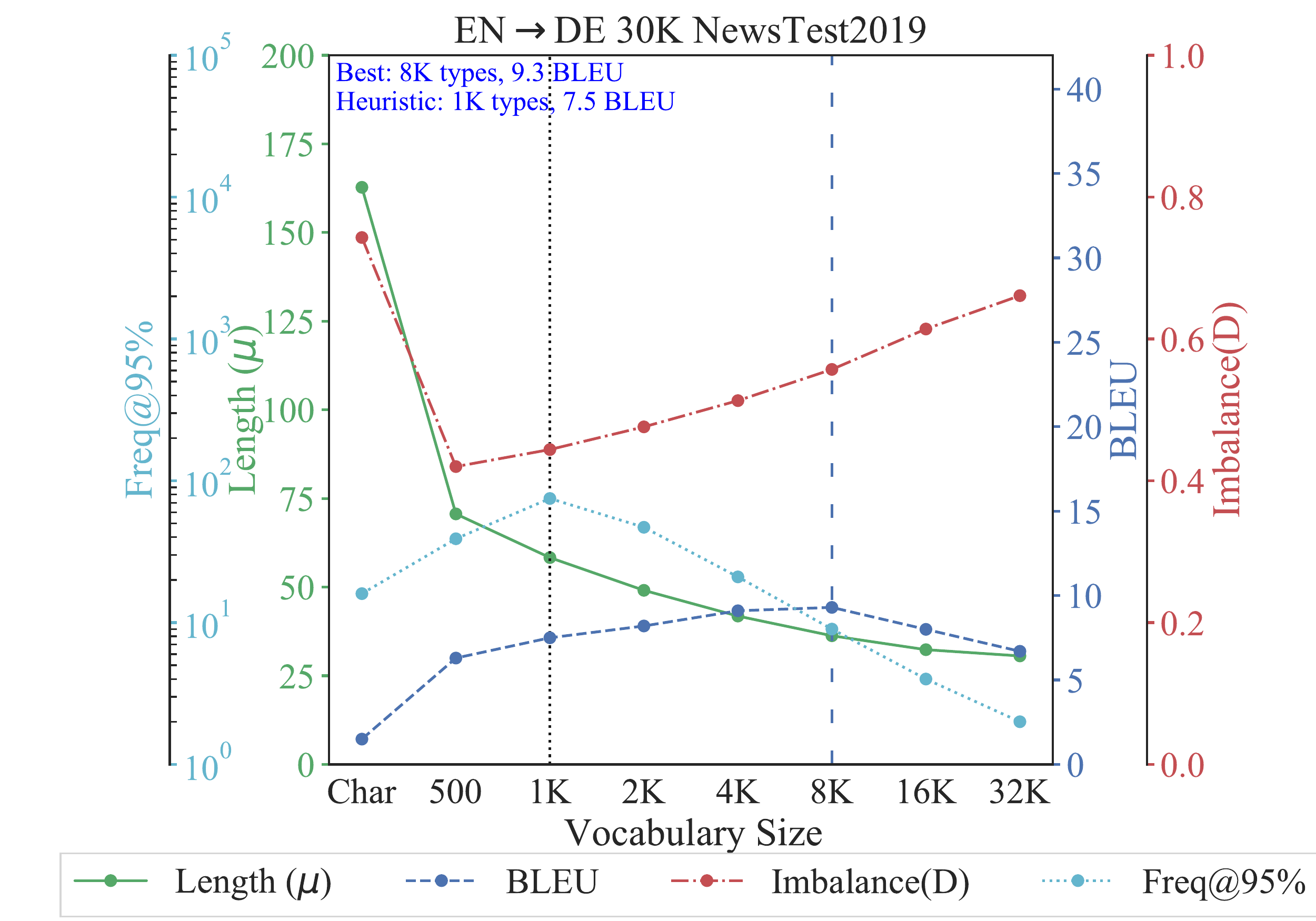}
\end{subfigure}
\begin{subfigure}{.32\textwidth}
  \centering
  \includegraphics[width=0.89\linewidth,trim={5.1cm 1.32cm 4.1cm 0},clip]{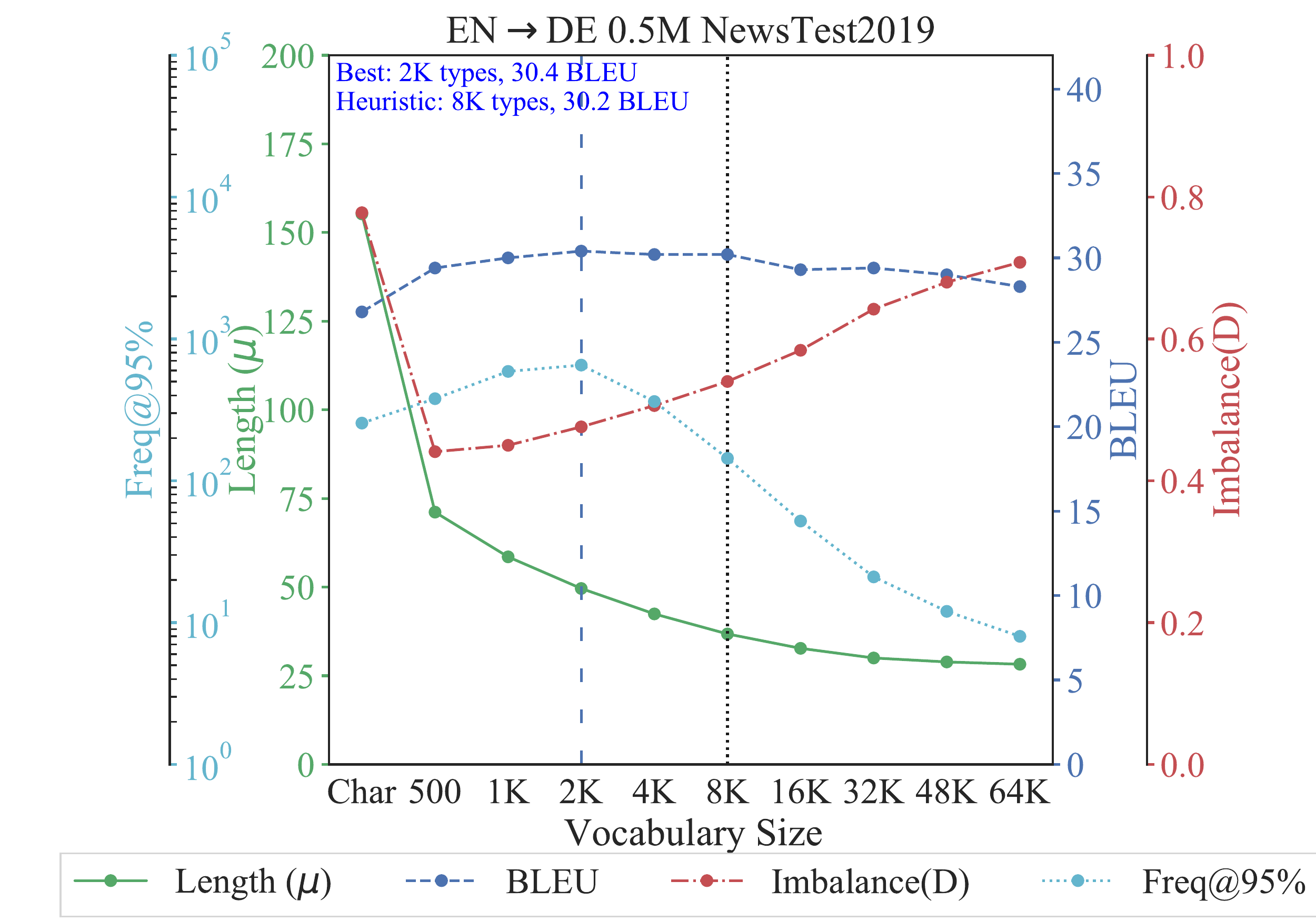}
\end{subfigure}
\begin{subfigure}{.33\textwidth}
  \centering
  \includegraphics[width=0.99\linewidth,trim={5.1cm 1.32cm 1.4cm 0},clip]{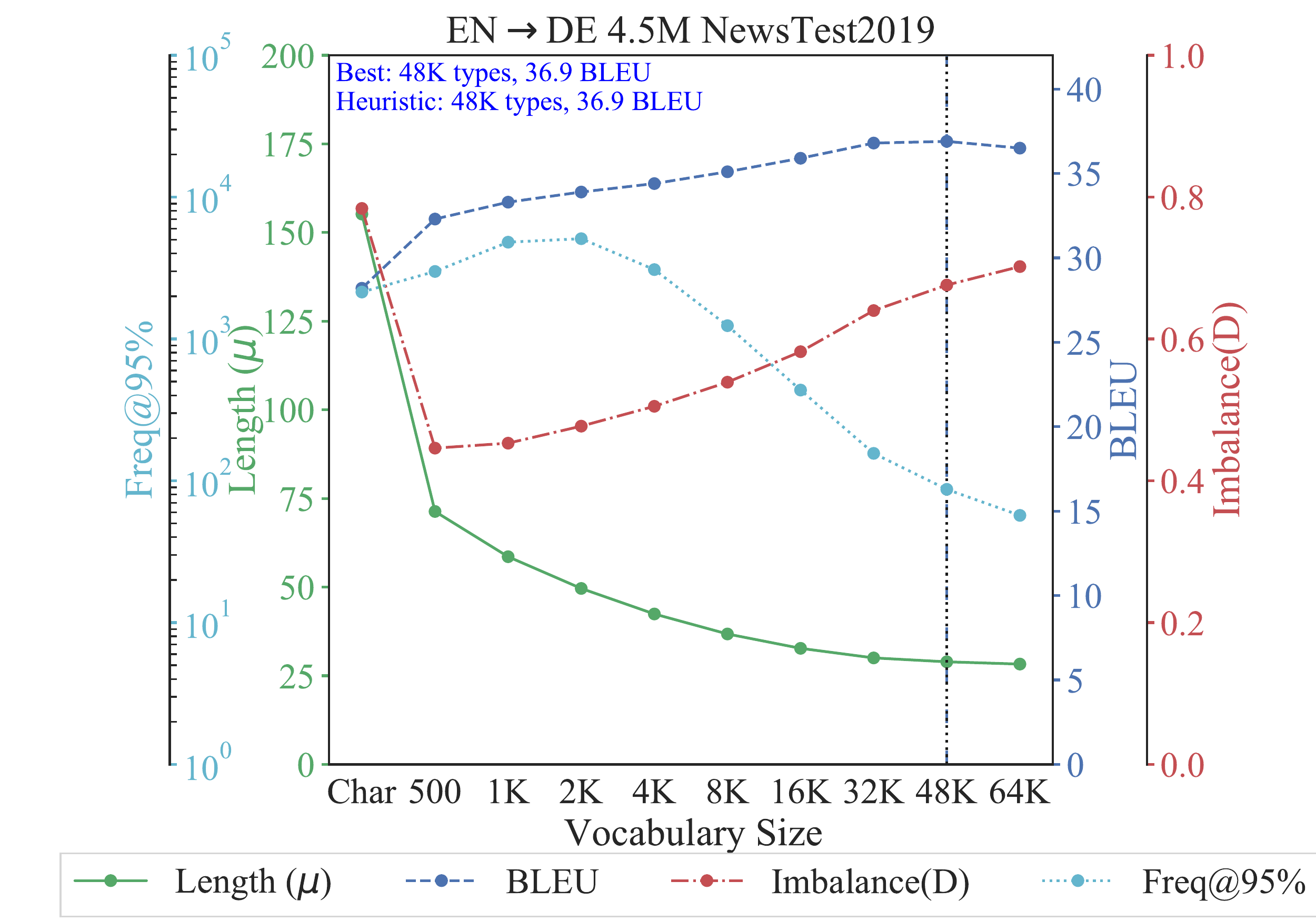}
\end{subfigure}

\begin{subfigure}{.33\textwidth}
  \centering
  \includegraphics[width=0.99\linewidth,trim={2.4cm 1.32cm 4.1cm 0},clip]{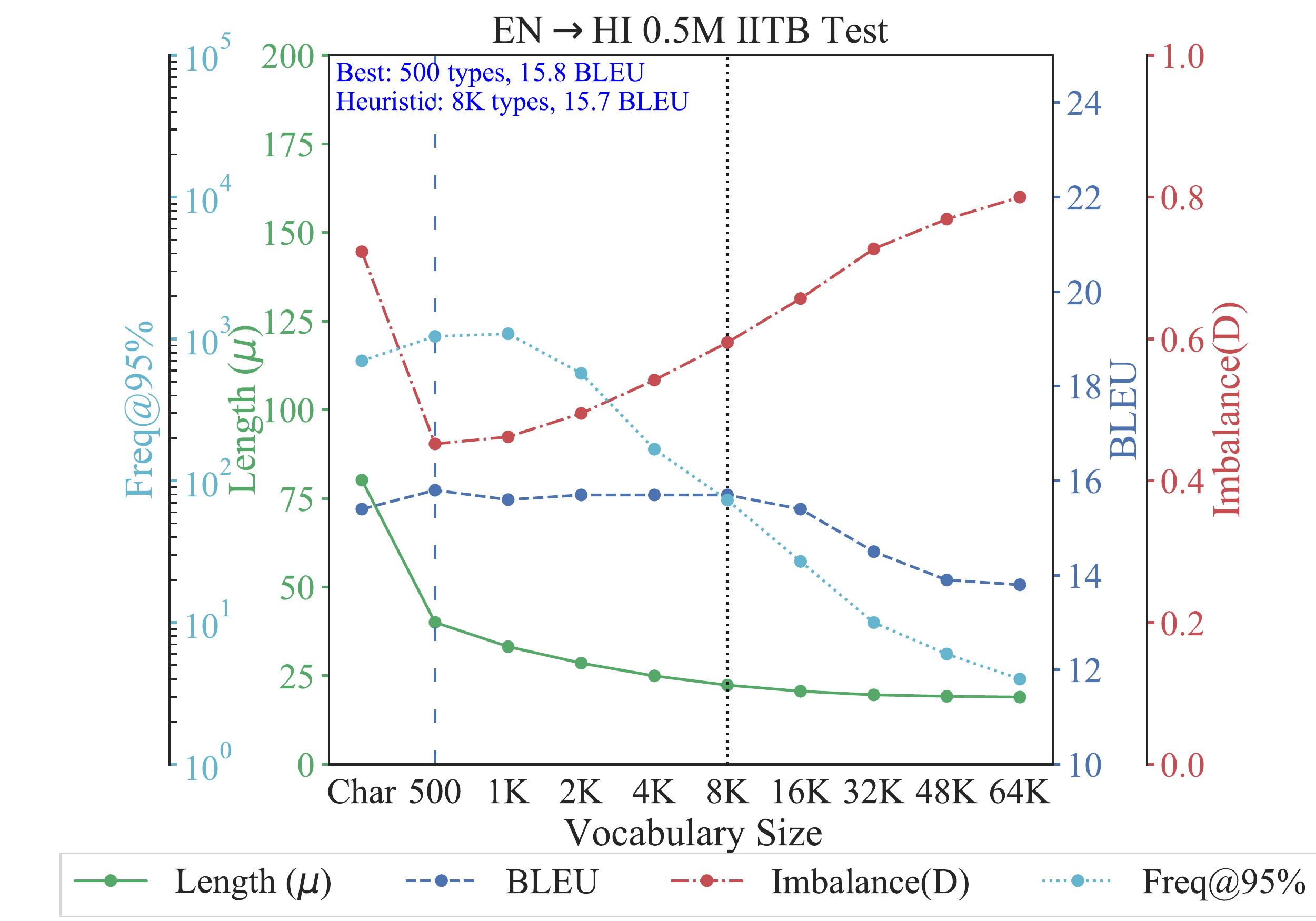}
\end{subfigure}
\begin{subfigure}{.32\textwidth}
  \centering
  \includegraphics[width=0.89\linewidth,trim={5.1cm 1.32cm 4.1cm 0},clip]{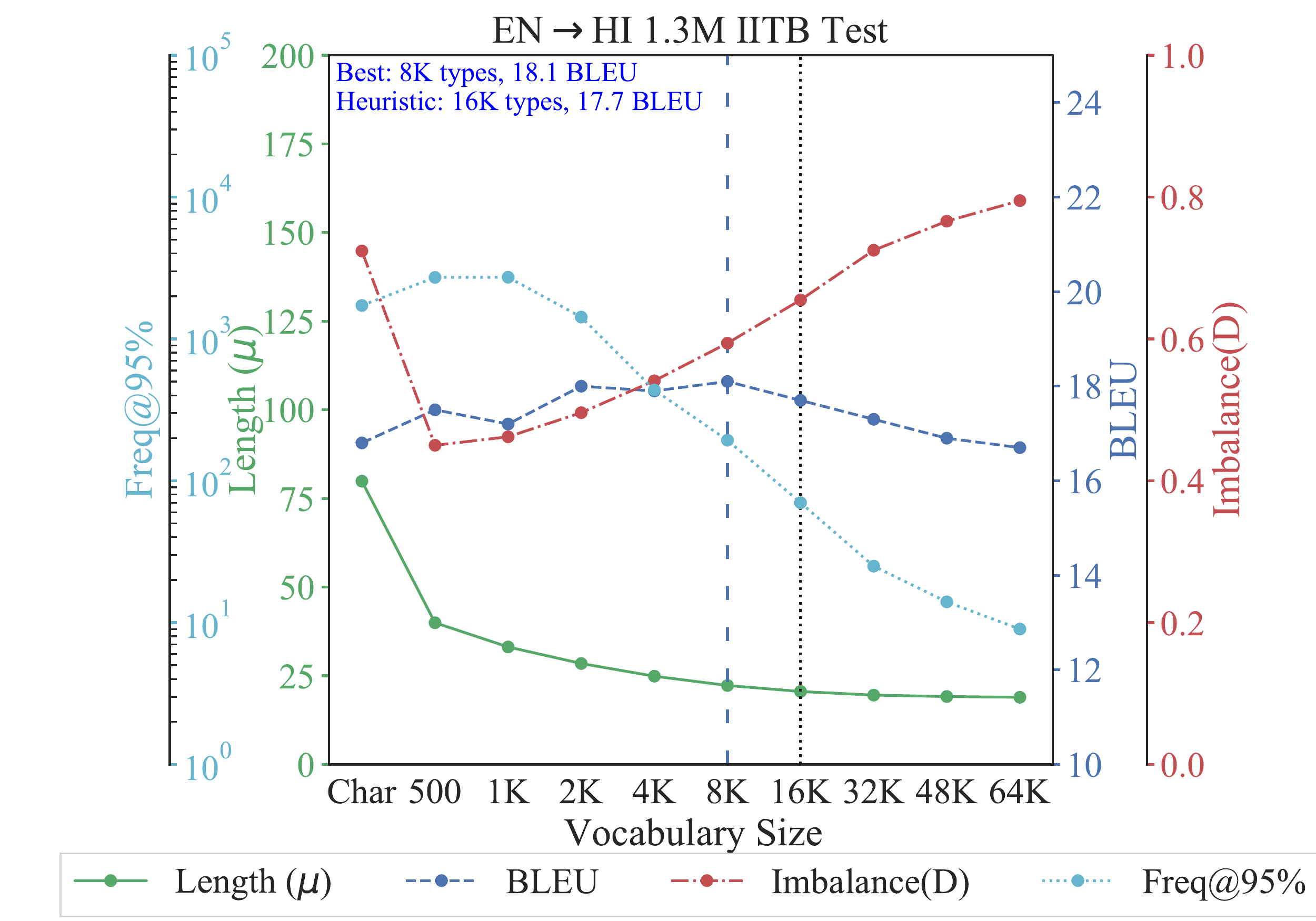}
\end{subfigure}
\begin{subfigure}{.33\textwidth}
  \centering
  \includegraphics[width=0.99\linewidth,trim={5.1cm 1.32cm 1.4cm 0},clip]{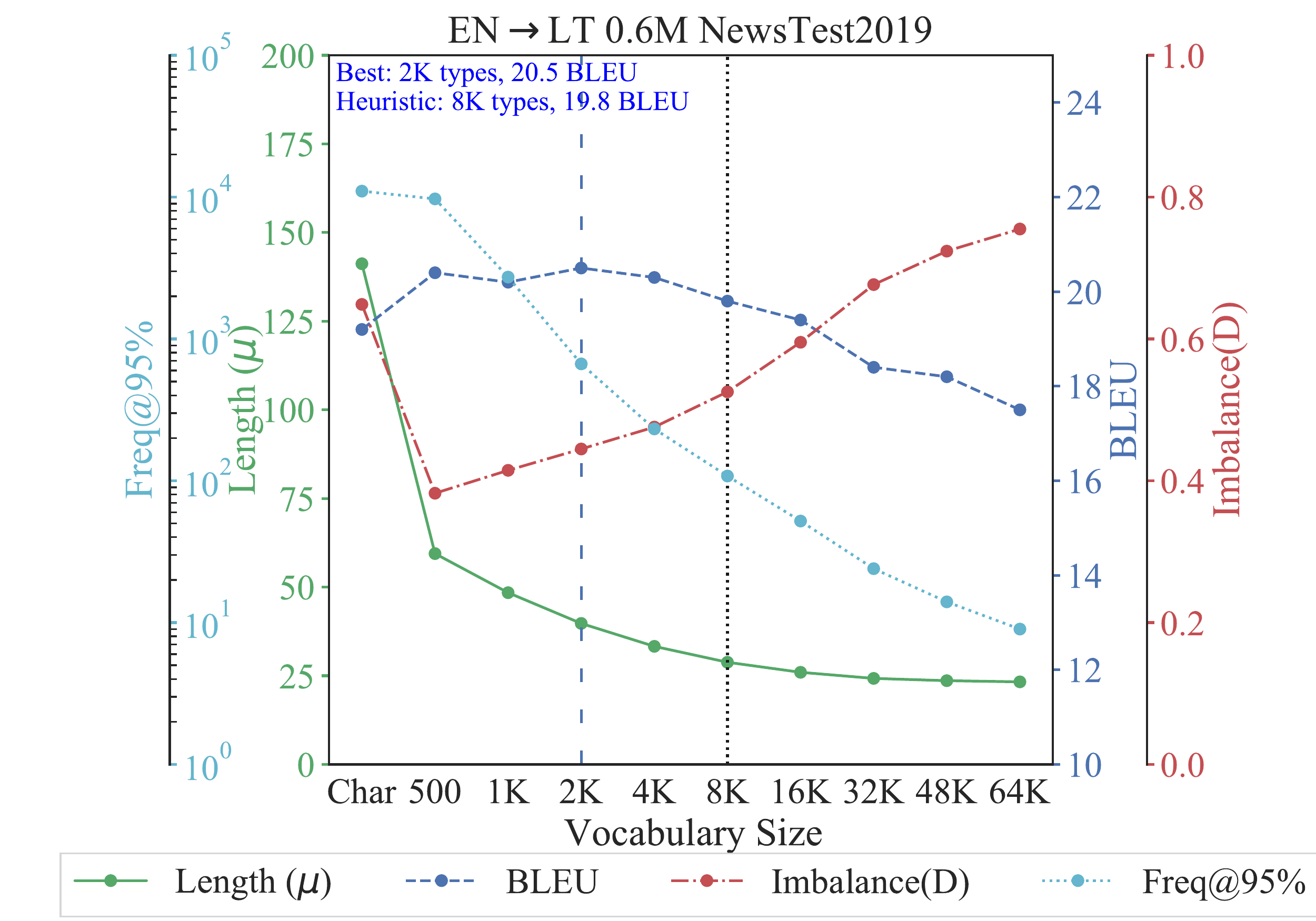}
\end{subfigure}

\caption{Visualization of sequence length ($\mu$) (lower is better), class imbalance (D) (lower is better), frequency of $95^{th}$ percentile class ($F_{95\%}$) (higher is better; plotted in logarithmic scale), and test set BLEU (higher is better) on all language pairs and training data sizes. DE$\leftrightarrow$EN of 1M resembles resembles DE$\leftrightarrow$EN of 0.5M and is provided in Appendix~\ref{sec:appendix} along with visualizations on validation sets.
The vocabulary sizes that achieved highest BLEU are indicated with dashed vertical lines, and the vocabulary our heuristic selects is indicated by dotted vertical lines.}
\label{fig:mu-d-freq-bleu}
\end{figure*}

Although Figures~\ref{fig:bleu-ende-deen} and \ref{fig:bleu-enhilt} indicate \textit{where} the optimal vocabulary size is for these chosen language pairs and datasets, the question of \textit{why} the peak is where it is remains unanswered.
We visualize $\mu$, $D$, and $F_{95\%}$ in Figure \ref{fig:mu-d-freq-bleu} to answer that question, and report these observations:
\begin{enumerate}
    \itemsep0em 
    \item Small vocabularies have a relatively larger $F_{95\%}$ (favorable to classifier), yet they are sub-optimal. We reason that this is due to the presence of a larger $\mu$, which is unfavorable to the autoregressor.
    \item Larger vocabularies such as 32K and beyond have a smaller $\mu$ which favors the autoregressor, yet rarely achieved the best BLEU.
    We reason this is due to the presence of a lower $F_{95\%}$ and a higher $D$ being unfavorable to the classifier.
    Since the larger datasets have many training examples for each class, as indicated by a generally larger $F_{95\%}$, we conclude that bigger vocabularies tend to yield optimal results compared to smaller datasets in the same language.
    
    \item On small (30K) to medium (1.3M) data sizes, the vocabulary size of 8K seems to find a good trade-off between $\mu$ and $D$, as well as between $\mu$ and $F_{95\%}$.
\end{enumerate}

 There is a \textit{simple heuristic} to locate the peak: the near-optimal vocabulary size is where sentence length $\mu$ is small, while $F_{95\%}$ is approximately $100$ or higher.

BLEU scores are often lower at larger vocabulary sizes---where $\mu$ is (favorably) low but $D$ is (unfavorably) high (Figure~\ref{fig:mu-d-freq-bleu}). This calls for a further investigation that is reported in the following section.

\section{Measuring Classifier Bias Due to Imbalance}
\label{sec:class-bias}

In a typical classification setting with imbalanced classes, the classifier learns an undesired bias based on frequencies. 

A balanced class distribution debiases in this regard, leading to improvement in the precision of frequent classes as well as recall of infrequent classes.
However, BLEU focuses only on the \textit{precision} of classes; except for adding a global brevity penalty, it is ignorant of the poor recall of infrequent classes. 

Therefore, the BLEU scores shown in Figures~\ref{fig:bleu-deen}, \ref{fig:bleu-ende} and \ref{fig:bleu-enhilt} capture only a part of the improvements and biases. 
In this section we perform a detailed analysis of the impact of class balancing by considering both precision \textit{and} recall of classes. 

We accomplish this in two stages:
First, we define a method to measure the bias of the model for classes based on their frequencies.
Second, we track the bias in relation to vocabulary size and class imbalance, and report DE$\rightarrow$EN, as it has many data points.

\subsection{Frequency Based Bias}
\begin{figure}[ht]
    \centering
    \includegraphics[width=\linewidth]{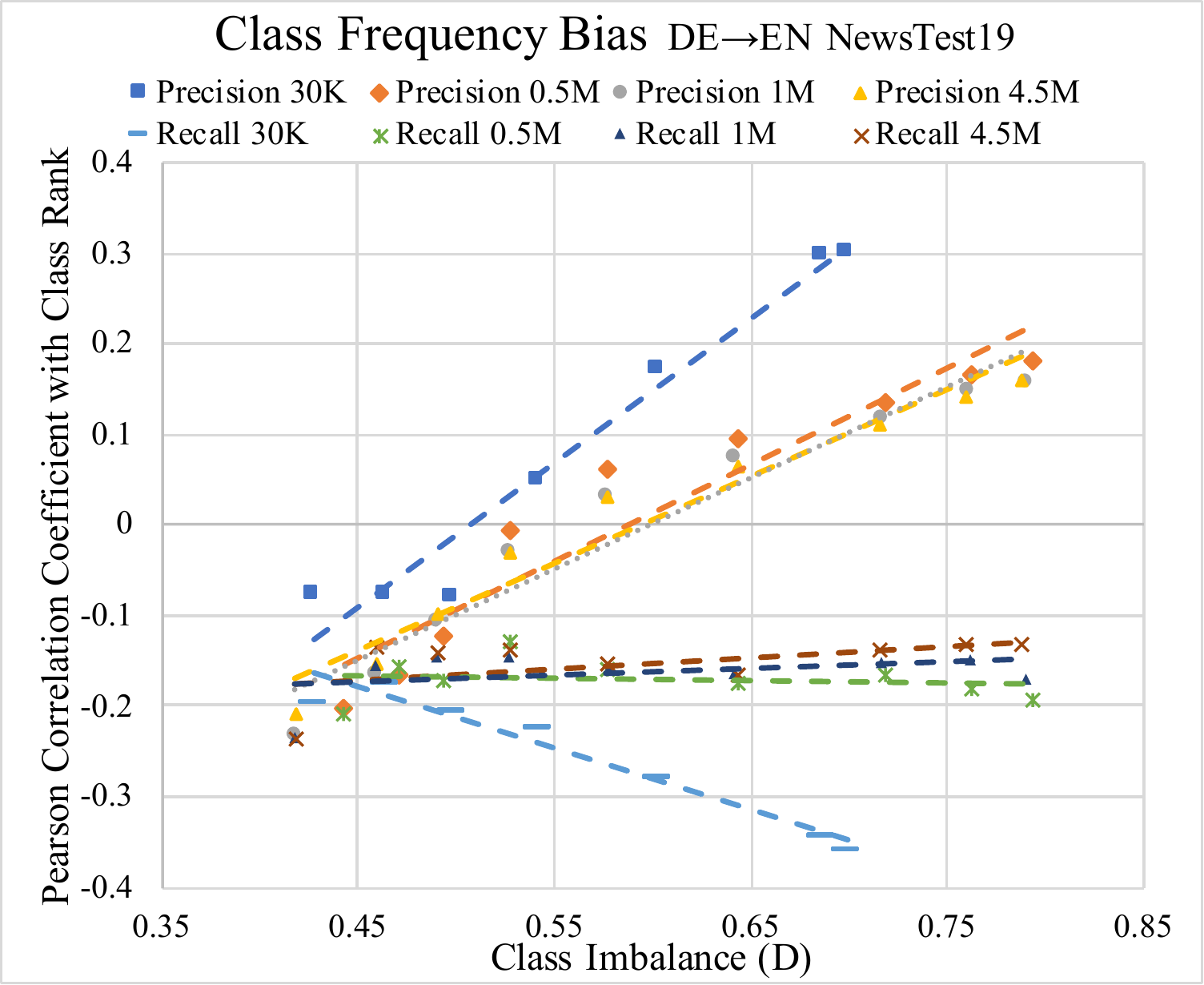}
        \caption{Correlation analysis on DE$\rightarrow$EN shows that NMT models suffer from frequency based class bias, indicated by non-zero correlation of both precision and recall with class rank.
        Reduction in class imbalance (D), as shown by the horizontal axis, generally reduces the bias as indicated by the reduction in magnitude of correlation.}
         \label{fig:corr-deen-test}
\end{figure}

We measure frequency bias using the Pearson correlation coefficient, $\rho$, between class rank and class performance, where for performance measures we use precision and recall.
Classes are ranked based on descending order of frequencies in the training data encoded with the same encoding schemes used for reported NMT experiments.
With this setup, the class with rank 1, say $F_1$, is the one with the highest frequency, rank 2 is the next highest, and so on.
More generally, $F_k$ is an index in the class rank list which has an inverse relation to class frequencies.
We define precision as $P_k$ for class $k$ similar to the unigram precision in BLEU and extend its definition to the unigram recall as $R_k$ (See Appendix \ref{apx:prec-recall-math} for detail).
The Pearson correlation coefficients between class rank and precision ($\rho_{F, P}$), and class rank and recall ($\rho_{F, R})$ are reported in Figure \ref{fig:corr-deen-test}.
In datasets where $D$ is high, the performance of classifier correlates with class rank. Such correlations are undesired for a classifier.

\subsection{Analysis of Class Frequency Bias}
An ideal classifier is one that does not discriminate classes based on their frequencies, i.e. one that exhibits no correlation between $\rho_{F, P}$, and$\rho_{F, R}$.  
However, we see in Figure~\ref{fig:corr-deen-test} that:
\begin{enumerate}
    \itemsep0em
    \item $\rho_{F, P}$ is positive when the dataset has high $D$; i.e if the class rank increases (frequency decreases), precision increases in relation to it.
    This indicates that frequent classes have relatively less precision than infrequent classes.
    The bias is strongly positive on smaller datasets such as 30K DE$\rightarrow$EN, which gradually diminishes if the training data size is increased or a vocabulary setting is chosen to reduce $D$.
    \item $\rho_{F, R}$ is negative, i.e., if the class rank increases, recall decreases in relation to it. 
    This is an indication that infrequent classes have relatively lower recall than frequent classes.
\end{enumerate}
Figure~\ref{fig:corr-deen-test} shows a trend that frequency based bias measured by correlation coefficient is lower in settings that have lower $D$.
However, since $D$ is non-zero, there still exists non-zero correlation between recall and class rank ($\rho_{F, R}$), indicating the poorer recall of low-frequency classes.

\section{Related Work}
\label{sec:related-work}

\subsection{NMT Architectures}
\label{sec:rel-nmt-arch}
Several variations of NMT models have been proposed and refined: \newcite{sutskever2014seq2seq} and \newcite{cho2014learning} introduce the RNN-based encoder-decoder model. 
\newcite{bahdanau2014nmtattn} introduce the attention mechanism and \newcite{luong2015effectiveAttn} propose several variations that became essential components of many future models.
RNN modules, either LSTM \cite{hochreiter1997LSTM} or GRU \cite{cho-etal-2014-properties}, have been popular choices for composing NMT encoders and decoders. 
The encoder uses bidirectional information, but the decoder is unidirectional, typically left-to-right, to facilitate autoregressive generation.
\newcite{gehring2017CNNMT} use a CNN architecture that outperforms RNN models.
\newcite{vaswani2017attention} propose the \textbf{Transformer}, whose main components are feed-forward and attention networks. 
There are only a few models that perform non-autoregressive NMT \cite{libovicky-helcl-2018-end,Gu-etal-17-NonAR-NMT}.
These are focused on improving the speed of inference; generation quality is currently sub-par compared to autoregressive models.
These non-autoregressive models can also be viewed as token classifiers with a different kind of feature extractor, whose strengths and limitations are yet to be theoretically understood.

\subsection{BPE Subwords} 
\newcite{sennrich-etal-2016-bpe} introduce BPE as a simplified way to solve out-of-vocabulary (OOV) words without having to use a back-off dictionary for OOV words.
They note that BPE improves the translation of not only the OOV words, but also some rare in-vocabulary words.
The analysis by \newcite{morishita-etal-2018-improving} is different than ours in that they view various vocabulary sizes as hierarchical features that are used in addition to a fixed vocabulary.
\newcite{DBLP:journals/corr/abs-1810-08641} offer an efficient way to search BPE vocabulary size for NMT.
\newcite{kudo-2018-subword} use BPE as a regularization technique by introducing sampling based randomness to the BPE segmentation. 
To the best of our knowledge, no previous work exists that analyzes BPE's effect on class imbalance.

\subsection{Class Imbalance}
\label{sec:rel-class-imb}
The class imbalance problem has been extensively studied in classical ML \cite{japkowicz2002ClassImbalance}.
In the medical domain \newcite{Maciej2008MedicalImbalance} find that classifier performance deteriorates with even modest imbalance in the training data.
Untreated class imbalance has been known to deteriorate the performance of image segmentation.  \newcite{Sudre2017GeneralizedDice} investigate the sensitivity of various loss functions.
\newcite{Johnson2019SurveyImbalance} survey imbalance learning and report that the effort is mostly targeted to computer vision tasks.
\newcite{buda-etal-2018-imbalance-cnn} provide a definition and quantification method for two types of class imbalance: \textit{step imbalance} and \textit{linear imbalance}.
Since the imbalance in Zipfian distribution of classes is neither single-stepped nor linear, we use a divergence based measure to quantify the imbalance.

\section{Conclusion}
\label{sec:conclusion}
Envisioning NMT as a token classifier with an autoregressor helps in analysing its weaknesses.
Our analysis provides an explanation of \textit{why} text generation using BPE vocabulary is more effective compared to word and character vocabularies, and \textit{why} certain BPE hyperparameters are better than others.
We show that the number of BPE merges is not an arbitrary hyperparameter, and that it can be tuned to address the class imbalance and sequence length problems. 
Our recommendation for Transformer NMT is to \textit{use the largest possible BPE vocabulary such that at least 95\% of classes have 100 or more examples in training}.
Even though certain BPE vocabulary sizes indirectly reduce the class imbalance, they do not completely eliminate it.
The class distributions after applying BPE contain sufficient imbalance for inducing the frequency based bias, especially affecting the recall of rare classes. 
Hence more effort in the future is needed to directly address the Zipfian imbalance.

\section*{Acknowledgments}

This research is based upon work supported in part by the Office of the Director of National Intelligence (ODNI), Intelligence Advanced Research Projects Activity (IARPA), via contract \# FA8650-17-C-9116, and by research sponsored by Air Force Research Laboratory (AFRL) under agreement number FA8750-19-1-1000. The views and conclusions contained herein are those of the authors and should not be interpreted as necessarily representing the official policies, either expressed or implied, of ODNI, IARPA, Air Force Laboratory, DARPA, or the U.S. Government. The U.S. Government is authorized to reproduce and distribute reprints for governmental purposes notwithstanding any copyright annotation therein.

\bibliography{nmt-class-imbalance}

\begin{thebibliography}{32}
\expandafter\ifx\csname natexlab\endcsname\relax\def\natexlab#1{#1}\fi

\bibitem[{Bahdanau et~al.(2015)Bahdanau, Cho, and Bengio}]{bahdanau2014nmtattn}
Dzmitry Bahdanau, Kyunghyun Cho, and Yoshua Bengio. 2015.
\newblock \href {http://arxiv.org/abs/1409.0473} {Neural machine translation by
  jointly learning to align and translate}.
\newblock In \emph{3rd International Conference on Learning Representations,
  {ICLR} 2015, San Diego, CA, USA, May 7-9, 2015, Conference Track
  Proceedings}.

\bibitem[{Barrault et~al.(2019)Barrault, Bojar, Costa-jussà, Federmann,
  Fishel, Graham, Haddow, Huck, Koehn, Malmasi, Monz, Müller, Pal, Post, and
  Zampieri}]{wmt19proceedings}
Loïc Barrault, Ondřej Bojar, Marta~R. Costa-jussà, Christian Federmann, Mark
  Fishel, Yvette Graham, Barry Haddow, Matthias Huck, Philipp Koehn, Shervin
  Malmasi, Christof Monz, Mathias Müller, Santanu Pal, Matt Post, and Marcos
  Zampieri. 2019.
\newblock \href {http://www.aclweb.org/anthology/W19-5301} {Findings of the
  2019 conference on machine translation ({WMT19})}.
\newblock In \emph{Proceedings of the Fourth Conference on Machine Translation
  (Volume 2: Shared Task Papers, Day 1)}, pages 1--61, Florence, Italy.
  Association for Computational Linguistics.

\bibitem[{Box et~al.(2015)Box, Jenkins, Reinsel, and Ljung}]{box2015time}
George~EP Box, Gwilym~M Jenkins, Gregory~C Reinsel, and Greta~M Ljung. 2015.
\newblock \emph{Time series analysis: forecasting and control}.
\newblock John Wiley \& Sons.

\bibitem[{Buda et~al.(2018)Buda, Maki, and
  Mazurowski}]{buda-etal-2018-imbalance-cnn}
Mateusz Buda, Atsuto Maki, and Maciej~A. Mazurowski. 2018.
\newblock \href {https://doi.org/https://doi.org/10.1016/j.neunet.2018.07.011}
  {A systematic study of the class imbalance problem in convolutional neural
  networks}.
\newblock \emph{Neural Networks}, 106:249 -- 259.

\bibitem[{Cho et~al.(2014{\natexlab{a}})Cho, van Merri{\"e}nboer, Bahdanau, and
  Bengio}]{cho-etal-2014-properties}
Kyunghyun Cho, Bart van Merri{\"e}nboer, Dzmitry Bahdanau, and Yoshua Bengio.
  2014{\natexlab{a}}.
\newblock \href {https://doi.org/10.3115/v1/W14-4012} {On the properties of
  neural machine translation: Encoder{--}decoder approaches}.
\newblock In \emph{Proceedings of {SSST}-8, Eighth Workshop on Syntax,
  Semantics and Structure in Statistical Translation}, pages 103--111, Doha,
  Qatar. Association for Computational Linguistics.

\bibitem[{Cho et~al.(2014{\natexlab{b}})Cho, van Merri{\"e}nboer, Gulcehre,
  Bahdanau, Bougares, Schwenk, and Bengio}]{cho2014learning}
Kyunghyun Cho, Bart van Merri{\"e}nboer, Caglar Gulcehre, Dzmitry Bahdanau,
  Fethi Bougares, Holger Schwenk, and Yoshua Bengio. 2014{\natexlab{b}}.
\newblock \href {https://doi.org/10.3115/v1/D14-1179} {Learning phrase
  representations using {RNN} encoder{--}decoder for statistical machine
  translation}.
\newblock In \emph{Proceedings of the 2014 Conference on Empirical Methods in
  Natural Language Processing ({EMNLP})}, pages 1724--1734, Doha, Qatar.
  Association for Computational Linguistics.

\bibitem[{Gehring et~al.(2017)Gehring, Auli, Grangier, Yarats, and
  Dauphin}]{gehring2017CNNMT}
Jonas Gehring, Michael Auli, David Grangier, Denis Yarats, and Yann~N Dauphin.
  2017.
\newblock Convolutional sequence to sequence learning.
\newblock In \emph{Proceedings of the 34th International Conference on Machine
  Learning-Volume 70}, pages 1243--1252. JMLR. org.

\bibitem[{Gu et~al.(2018)Gu, Bradbury, Xiong, Li, and
  Socher}]{Gu-etal-17-NonAR-NMT}
Jiatao Gu, James Bradbury, Caiming Xiong, Victor O.~K. Li, and Richard Socher.
  2018.
\newblock \href {https://openreview.net/forum?id=B1l8BtlCb} {Non-autoregressive
  neural machine translation}.
\newblock In \emph{6th International Conference on Learning Representations,
  {ICLR} 2018, Vancouver, BC, Canada, April 30 - May 3, 2018, Conference Track
  Proceedings}. OpenReview.net.

\bibitem[{Hochreiter and Schmidhuber(1997)}]{hochreiter1997LSTM}
Sepp Hochreiter and J{\"u}rgen Schmidhuber. 1997.
\newblock Long short-term memory.
\newblock \emph{Neural computation}, 9(8):1735--1780.

\bibitem[{Japkowicz and Stephen(2002)}]{japkowicz2002ClassImbalance}
Nathalie Japkowicz and Shaju Stephen. 2002.
\newblock The class imbalance problem: A systematic study.
\newblock \emph{Intelligent Data Analysis}, 6(5):429--449.

\bibitem[{Johnson and Khoshgoftaar(2019)}]{Johnson2019SurveyImbalance}
Justin~M. Johnson and Taghi~M. Khoshgoftaar. 2019.
\newblock \href {https://doi.org/10.1186/s40537-019-0192-5} {Survey on deep
  learning with class imbalance}.
\newblock \emph{Journal of Big Data}, 6(1):27.

\bibitem[{Kingma and Ba(2015)}]{kingma2015adam}
Diederik~P. Kingma and Jimmy Ba. 2015.
\newblock \href {http://arxiv.org/abs/1412.6980} {Adam: {A} method for
  stochastic optimization}.
\newblock In \emph{3rd International Conference on Learning Representations,
  {ICLR} 2015, San Diego, CA, USA, May 7-9, 2015, Conference Track
  Proceedings}.

\bibitem[{Koehn and Knowles(2017)}]{koehn2017sixchallenges}
Philipp Koehn and Rebecca Knowles. 2017.
\newblock \href {https://doi.org/10.18653/v1/W17-3204} {Six challenges for
  neural machine translation}.
\newblock In \emph{Proceedings of the First Workshop on Neural Machine
  Translation}, pages 28--39, Vancouver. Association for Computational
  Linguistics.

\bibitem[{Kudo(2018)}]{kudo-2018-subword}
Taku Kudo. 2018.
\newblock \href {https://doi.org/10.18653/v1/P18-1007} {Subword regularization:
  Improving neural network translation models with multiple subword
  candidates}.
\newblock In \emph{Proceedings of the 56th Annual Meeting of the Association
  for Computational Linguistics (Volume 1: Long Papers)}, pages 66--75,
  Melbourne, Australia. Association for Computational Linguistics.

\bibitem[{Kunchukuttan et~al.(2018)Kunchukuttan, Mehta, and
  Bhattacharyya}]{iitb-hien}
Anoop Kunchukuttan, Pratik Mehta, and Pushpak Bhattacharyya. 2018.
\newblock \href {https://www.aclweb.org/anthology/L18-1548} {The {IIT} {B}ombay
  {E}nglish-{H}indi parallel corpus}.
\newblock In \emph{Proceedings of the Eleventh International Conference on
  Language Resources and Evaluation ({LREC} 2018)}, Miyazaki, Japan. European
  Language Resources Association (ELRA).

\bibitem[{Libovick{\'y} and Helcl(2018)}]{libovicky-helcl-2018-end}
Jind{\v{r}}ich Libovick{\'y} and Jind{\v{r}}ich Helcl. 2018.
\newblock \href {https://doi.org/10.18653/v1/D18-1336} {End-to-end
  non-autoregressive neural machine translation with connectionist temporal
  classification}.
\newblock In \emph{Proceedings of the 2018 Conference on Empirical Methods in
  Natural Language Processing}, pages 3016--3021, Brussels, Belgium.
  Association for Computational Linguistics.

\bibitem[{Luong et~al.(2015)Luong, Pham, and Manning}]{luong2015effectiveAttn}
Thang Luong, Hieu Pham, and Christopher~D. Manning. 2015.
\newblock \href {https://doi.org/10.18653/v1/D15-1166} {Effective approaches to
  attention-based neural machine translation}.
\newblock In \emph{Proceedings of the 2015 Conference on Empirical Methods in
  Natural Language Processing}, pages 1412--1421, Lisbon, Portugal. Association
  for Computational Linguistics.

\bibitem[{Maas et~al.(2011)Maas, Daly, Pham, Huang, Ng, and
  Potts}]{maas-etal-2011-learning}
Andrew~L. Maas, Raymond~E. Daly, Peter~T. Pham, Dan Huang, Andrew~Y. Ng, and
  Christopher Potts. 2011.
\newblock \href {https://www.aclweb.org/anthology/P11-1015} {Learning word
  vectors for sentiment analysis}.
\newblock In \emph{Proceedings of the 49th Annual Meeting of the Association
  for Computational Linguistics: Human Language Technologies}, pages 142--150,
  Portland, Oregon, USA. Association for Computational Linguistics.

\bibitem[{Mazurowski et~al.(2008)Mazurowski, Habas, Zurada, Lo, Baker, and
  Tourassi}]{Maciej2008MedicalImbalance}
Maciej~A. Mazurowski, Piotr~A. Habas, Jacek~M. Zurada, Joseph~Y. Lo, Jay~A.
  Baker, and Georgia~D. Tourassi. 2008.
\newblock \href {https://doi.org/https://doi.org/10.1016/j.neunet.2007.12.031}
  {Training neural network classifiers for medical decision making: The effects
  of imbalanced datasets on classification performance}.
\newblock \emph{Neural Networks}, 21(2):427 -- 436.
\newblock Advances in Neural Networks Research: IJCNN ’07.

\bibitem[{Morishita et~al.(2018)Morishita, Suzuki, and
  Nagata}]{morishita-etal-2018-improving}
Makoto Morishita, Jun Suzuki, and Masaaki Nagata. 2018.
\newblock \href {https://www.aclweb.org/anthology/C18-1052} {Improving neural
  machine translation by incorporating hierarchical subword features}.
\newblock In \emph{Proceedings of the 27th International Conference on
  Computational Linguistics}, pages 618--629, Santa Fe, New Mexico, USA.
  Association for Computational Linguistics.

\bibitem[{Popel and Bojar(2018)}]{popel2018tfm-train-tips}
Martin Popel and Ond{\v{r}}ej Bojar. 2018.
\newblock Training tips for the transformer model.
\newblock \emph{The Prague Bulletin of Mathematical Linguistics},
  110(1):43--70.

\bibitem[{Post(2018)}]{post-2018-sacreBLEU}
Matt Post. 2018.
\newblock \href {https://www.aclweb.org/anthology/W18-6319} {A call for clarity
  in reporting {BLEU} scores}.
\newblock In \emph{Proceedings of the Third Conference on Machine Translation:
  Research Papers}, pages 186--191, Belgium, Brussels. Association for
  Computational Linguistics.

\bibitem[{Press and Wolf(2017)}]{press-wolf-2017-using}
Ofir Press and Lior Wolf. 2017.
\newblock \href {https://www.aclweb.org/anthology/E17-2025} {Using the output
  embedding to improve language models}.
\newblock In \emph{Proceedings of the 15th Conference of the {E}uropean Chapter
  of the Association for Computational Linguistics: Volume 2, Short Papers},
  pages 157--163, Valencia, Spain. Association for Computational Linguistics.

\bibitem[{Salesky et~al.(2018)Salesky, Runge, Coda, Niehues, and
  Neubig}]{DBLP:journals/corr/abs-1810-08641}
Elizabeth Salesky, Andrew Runge, Alex Coda, Jan Niehues, and Graham Neubig.
  2018.
\newblock \href {http://arxiv.org/abs/1810.08641} {Optimizing segmentation
  granularity for neural machine translation}.
\newblock \emph{CoRR}, abs/1810.08641.

\bibitem[{Sennrich et~al.(2016)Sennrich, Haddow, and
  Birch}]{sennrich-etal-2016-bpe}
Rico Sennrich, Barry Haddow, and Alexandra Birch. 2016.
\newblock \href {https://doi.org/10.18653/v1/P16-1162} {Neural machine
  translation of rare words with subword units}.
\newblock In \emph{Proceedings of the 54th Annual Meeting of the Association
  for Computational Linguistics (Volume 1: Long Papers)}, pages 1715--1725,
  Berlin, Germany. Association for Computational Linguistics.

\bibitem[{Steedman(2008)}]{steedman-2008-last}
Mark Steedman. 2008.
\newblock \href {https://doi.org/10.1162/coli.2008.34.1.137} {On becoming a
  discipline}.
\newblock \emph{Computational Linguistics}, 34(1):137--144.

\bibitem[{Sudre et~al.(2017)Sudre, Li, Vercauteren, Ourselin, and
  Jorge~Cardoso}]{Sudre2017GeneralizedDice}
Carole~H. Sudre, Wenqi Li, Tom Vercauteren, Sebastien Ourselin, and
  M.~Jorge~Cardoso. 2017.
\newblock Generalised dice overlap as a deep learning loss function for highly
  unbalanced segmentations.
\newblock In \emph{Deep Learning in Medical Image Analysis and Multimodal
  Learning for Clinical Decision Support}, pages 240--248, Cham. Springer
  International Publishing.

\bibitem[{Sutskever et~al.(2014)Sutskever, Vinyals, and
  Le}]{sutskever2014seq2seq}
Ilya Sutskever, Oriol Vinyals, and Quoc~V Le. 2014.
\newblock Sequence to sequence learning with neural networks.
\newblock In \emph{Advances in neural information processing systems}, pages
  3104--3112.

\bibitem[{Tjong Kim~Sang and De~Meulder(2003)}]{CoNLL-2003-NER}
Erik~F. Tjong Kim~Sang and Fien De~Meulder. 2003.
\newblock \href {https://www.aclweb.org/anthology/W03-0419} {Introduction to
  the {C}o{NLL}-2003 shared task: Language-independent named entity
  recognition}.
\newblock In \emph{Proceedings of the Seventh Conference on Natural Language
  Learning at {HLT}-{NAACL} 2003}, pages 142--147.

\bibitem[{Vaswani et~al.(2017)Vaswani, Shazeer, Parmar, Uszkoreit, Jones,
  Gomez, Kaiser, and Polosukhin}]{vaswani2017attention}
Ashish Vaswani, Noam Shazeer, Niki Parmar, Jakob Uszkoreit, Llion Jones,
  Aidan~N Gomez, {\L}ukasz Kaiser, and Illia Polosukhin. 2017.
\newblock Attention is all you need.
\newblock In \emph{Advances in Neural Information Processing Systems}, pages
  5998--6008.

\bibitem[{Zeman et~al.(2017)Zeman, Popel, Straka, Hajic, Nivre, Ginter,
  Luotolahti, Pyysalo, Petrov, Potthast, Tyers, Badmaeva, Gokirmak, Nedoluzhko,
  Cinkova, Hajic~jr., Hlavacova, Kettnerov\'{a}, Uresova, Kanerva, Ojala,
  Missil\"{a}, Manning, Schuster, Reddy, Taji, Habash, Leung, de~Marneffe,
  Sanguinetti, Simi, Kanayama, dePaiva, Droganova, Mart\'{i}nez~Alonso,
  \c{C}\"{o}ltekin, Sulubacak, Uszkoreit, Macketanz, Burchardt, Harris,
  Marheinecke, Rehm, Kayadelen, Attia, Elkahky, Yu, Pitler, Lertpradit, Mandl,
  Kirchner, Alcalde, Strnadov\'{a}, Banerjee, Manurung, Stella, Shimada, Kwak,
  Mendonca, Lando, Nitisaroj, and Li}]{CoNLL2017-shared-UD}
Daniel Zeman, Martin Popel, Milan Straka, Jan Hajic, Joakim Nivre, Filip
  Ginter, Juhani Luotolahti, Sampo Pyysalo, Slav Petrov, Martin Potthast,
  Francis Tyers, Elena Badmaeva, Memduh Gokirmak, Anna Nedoluzhko, Silvie
  Cinkova, Jan Hajic~jr., Jaroslava Hlavacova, V\'{a}clava Kettnerov\'{a},
  Zdenka Uresova, Jenna Kanerva, Stina Ojala, Anna Missil\"{a}, Christopher~D.
  Manning, Sebastian Schuster, Siva Reddy, Dima Taji, Nizar Habash, Herman
  Leung, Marie-Catherine de~Marneffe, Manuela Sanguinetti, Maria Simi, Hiroshi
  Kanayama, Valeria dePaiva, Kira Droganova, H\'{e}ctor Mart\'{i}nez~Alonso,
  \c{C}a\u{g}rı \c{C}\"{o}ltekin, Umut Sulubacak, Hans Uszkoreit, Vivien
  Macketanz, Aljoscha Burchardt, Kim Harris, Katrin Marheinecke, Georg Rehm,
  Tolga Kayadelen, Mohammed Attia, Ali Elkahky, Zhuoran Yu, Emily Pitler, Saran
  Lertpradit, Michael Mandl, Jesse Kirchner, Hector~Fernandez Alcalde, Jana
  Strnadov\'{a}, Esha Banerjee, Ruli Manurung, Antonio Stella, Atsuko Shimada,
  Sookyoung Kwak, Gustavo Mendonca, Tatiana Lando, Rattima Nitisaroj, and Josie
  Li. 2017.
\newblock \href {http://www.aclweb.org/anthology/K/K17/K17-3001.pdf} {Conll
  2017 shared task: Multilingual parsing from raw text to universal
  dependencies}.
\newblock In \emph{Proceedings of the CoNLL 2017 Shared Task: Multilingual
  Parsing from Raw Text to Universal Dependencies}, pages 1--19, Vancouver,
  Canada. Association for Computational Linguistics.

\bibitem[{Zhang et~al.(2015)Zhang, Zhao, and
  LeCun}]{Zhang-etal-15-cnn-sentiment}
Xiang Zhang, Junbo Zhao, and Yann LeCun. 2015.
\newblock \href {http://dl.acm.org/citation.cfm?id=2969239.2969312}
  {Character-level convolutional networks for text classification}.
\newblock In \emph{Proceedings of the 28th International Conference on Neural
  Information Processing Systems - Volume 1}, NIPS'15, pages 649--657,
  Cambridge, MA, USA. MIT Press.

\end{thebibliography}
\bibliographystyle{emnlp-2020-template/acl_natbib}

\newpage

\appendix

\section{Unigram Precision and Recall}
\label{apx:prec-recall-math}
For the sake of clarity, consider a test dataset $T$ of $N$ pairs of parallel sentences, $(x^{(i)}, y^{(i)})$ where $x^{(i)}$ and $y^{(i)}$ are the $i^{th}$ source and reference sequences, respectively. 
We use single reference $y^{(i)}$ translations for this analysis. 
For each $x^{(i)}$, let $h^{(i)}$ be the translation hypothesis from an MT model.

Let the indicator $\mathbbm{1}_k^{a}$ have value $1$ iff type $c_k$ exists in sequence $a$, where $a$ can be either hypothesis $h^{(i)}$ or reference $y^{(i)}$. 
The function $count(c_k, a)$ counts the times token $c_k$ exists in sequence $a$; $match(c_k, y^{(i)}, h^{(i)})$ returns the times $c_k$ is matched between hypothesis and reference, given by $min\{count(c_k, y^{(i)}), count(c_k, h^{(i)})\}$ 

Let $P_k^{(i)}$ and $R_k^{(i)}$ be precision and recall of $c_k$ on a specific record $i \in T$, given by:
$$ P_k^{(i)} = \frac{match(c_k, y^{(i)}, h^{(i)})} {count(c_k, h^{(i)})} \text{, defined iff } \mathbbm{1}_k^{h^{(i)}} $$ 
$$ R_k^{(i)} = \frac{match(c_k, y^{(i)}, h^{(i)})}{count(c_k, y^{(i)})} \text{, defined iff } \mathbbm{1}_k^{y^{(i)}} $$

Let $P_k$, $R_k$ be the expected precision and recall for $c_k$ over the whole $T$, given by:
$$P_k = \mathbbm{E}_{i\in T}[P_k^{(i)}] = \frac{\sum_{i=1}^{N} \mathbbm{1}_k^{h^{(i)}} P_k^{(i)}} {\sum_{i=1}^{N} \mathbbm{1}_k^{h^{(i)}}}$$ 
$$R_k = \mathbbm{E}_{i \in T}[R_k^{(i)}] = \frac{\sum_{i=1}^{N} \mathbbm{1}_k^{y^{(i)}} R_k^{(i)}} {\sum_{i=1}^{N} \mathbbm{1}_k^{y^{(i)}} }$$ 

\section{More Visualizations}
\label{sec:appendix}

\begin{figure}[htp]

\subfloat[]{%
  \includegraphics[clip,width=\columnwidth]{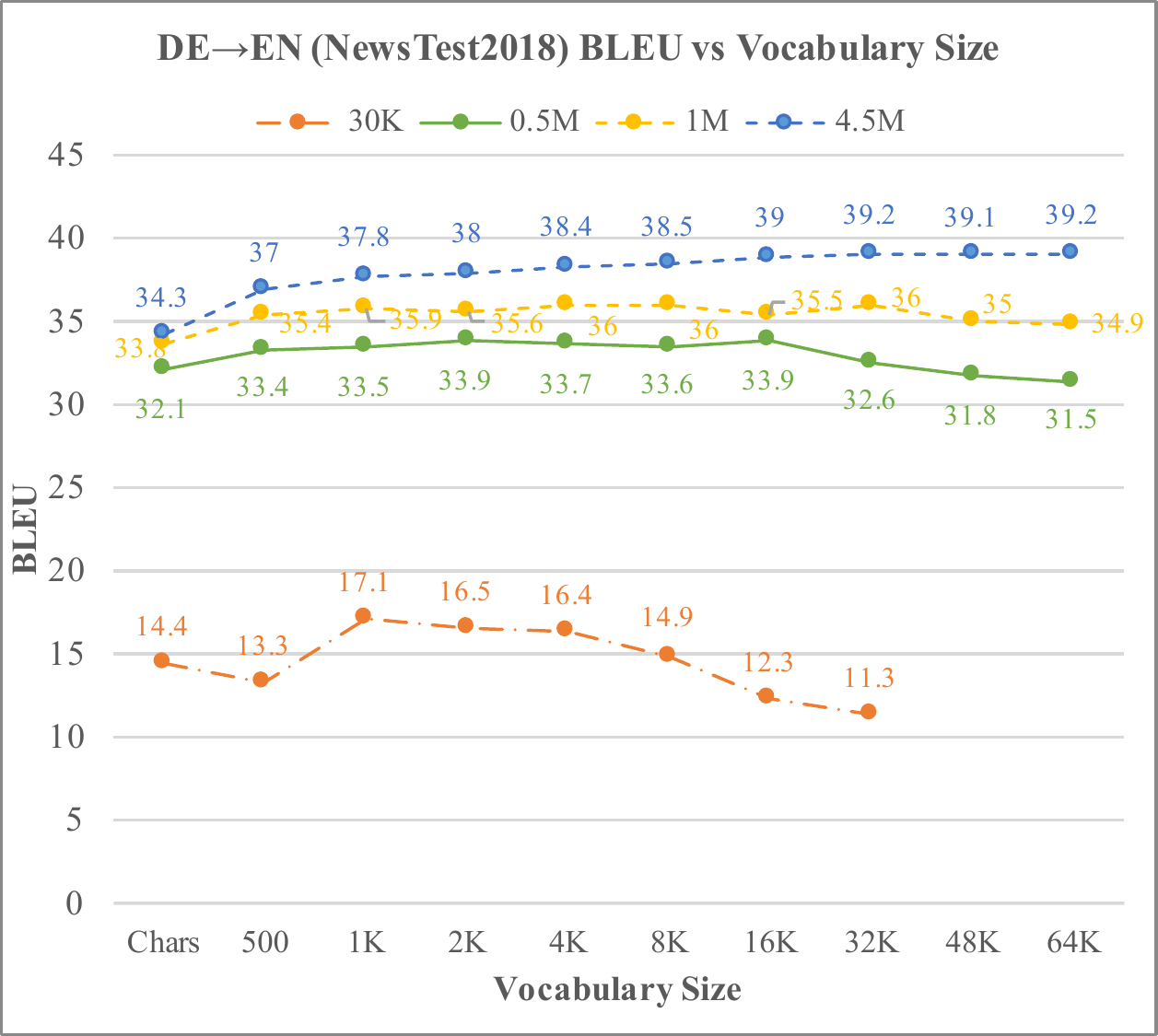}%
  
}

\subfloat[]{%
  \includegraphics[clip,width=\columnwidth]{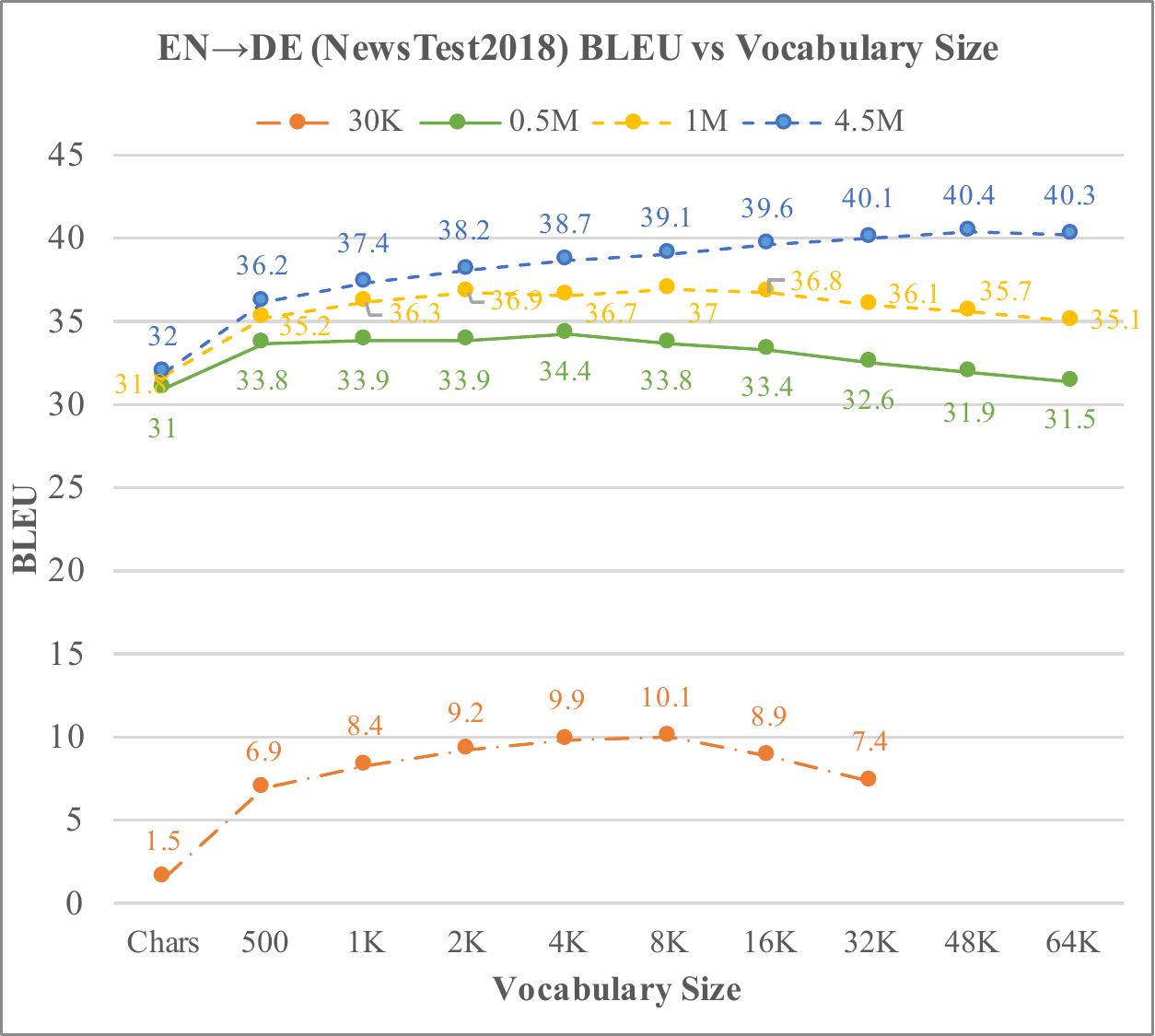}%
}

\subfloat[]{%
  \includegraphics[clip,width=\columnwidth]{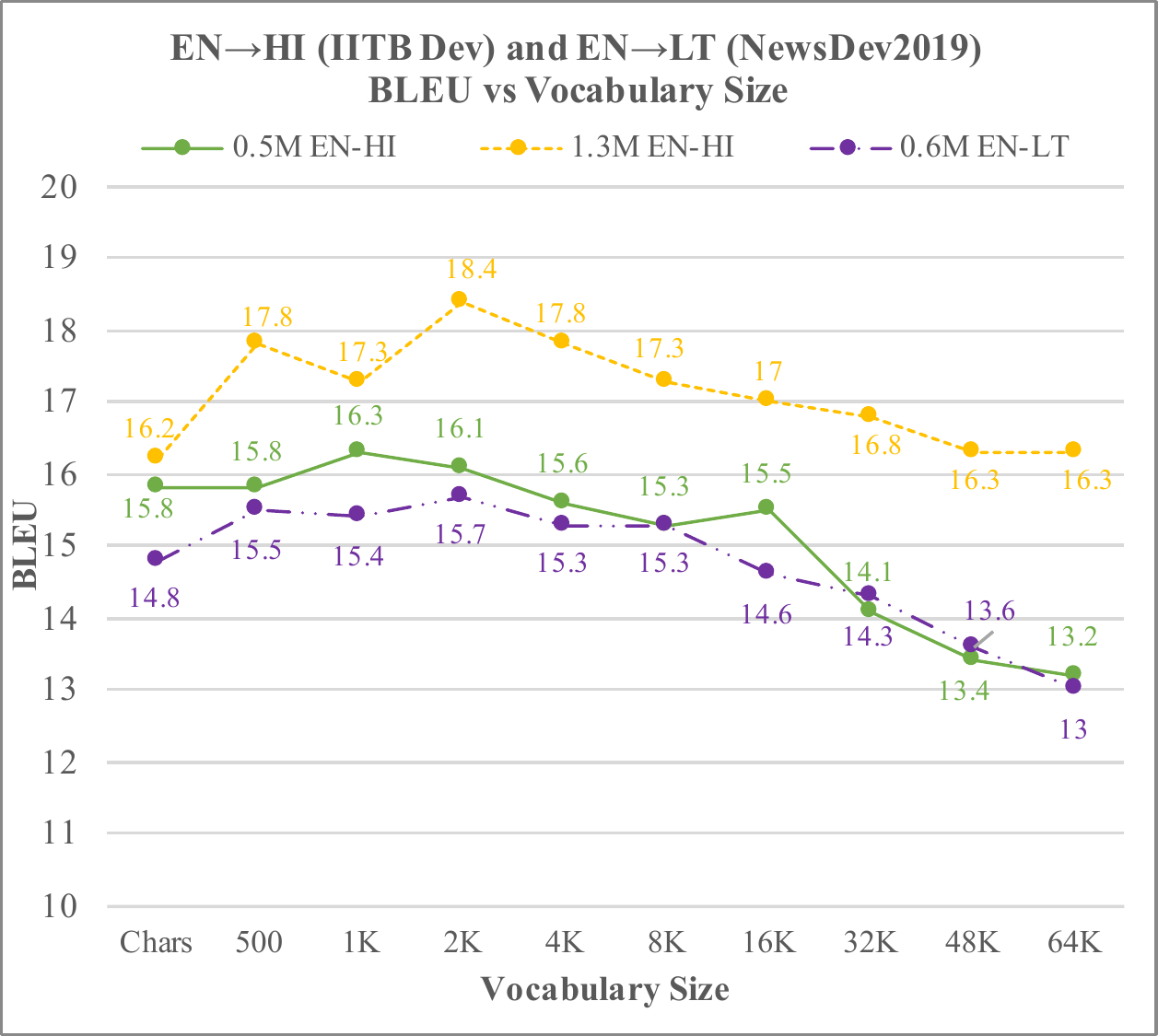}%
}

\caption{BLEU on all validation sets as a function of vocabulary size at various training set sizes.}
\label{fig:bleu-all-dev}

\end{figure}

We provide BLEU scores on validation datasets in Figure  ~\ref{fig:bleu-all-dev}.


\begin{figure}[ht]
\begin{subfigure}{\linewidth}
  \centering
  \includegraphics[width=\linewidth,trim={1.4cm 0 0.2cm 16.45cm},clip]{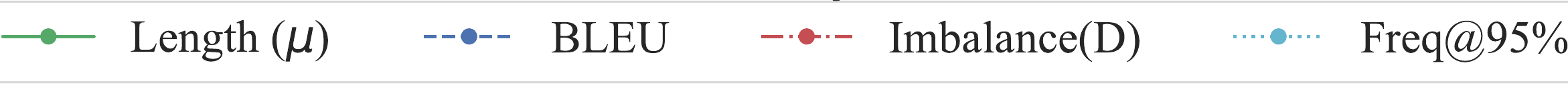}
\end{subfigure}

\begin{subfigure}{\linewidth}
  \centering
  \includegraphics[width=0.99\linewidth,trim={2.4cm 1.32cm 1.4cm 0},clip]{4axv-test-deen-1m.pdf}
\end{subfigure}

\begin{subfigure}{\linewidth}
  \centering
  \includegraphics[width=\linewidth,trim={2.4cm 1.32cm 1.4cm 0},clip]{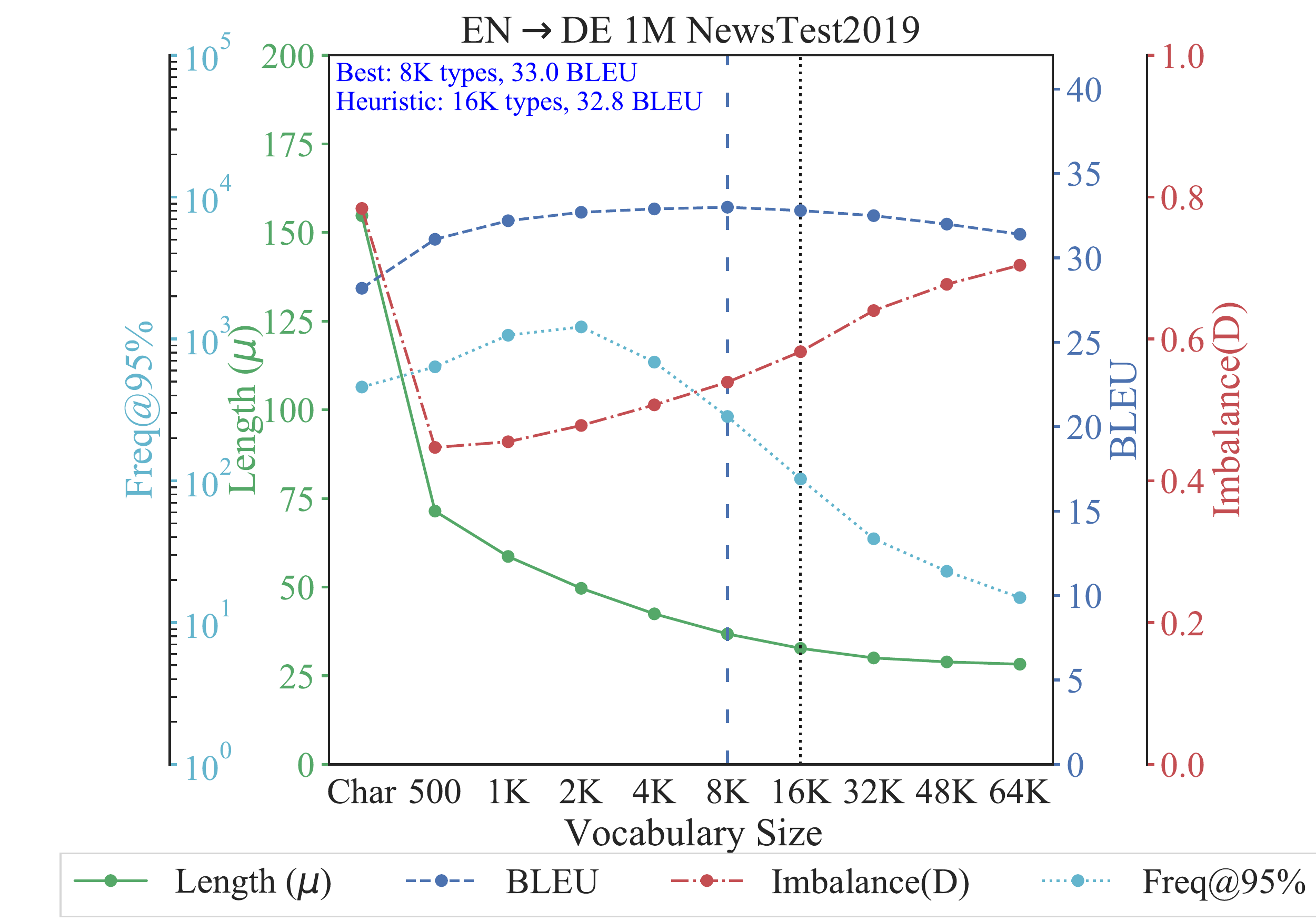}
\end{subfigure}

\caption{Visualization of sequence length ($\mu$) (lower is better), class imbalance (D) (lower is better), frequency of $95^{th}$ percentile class ($F_{95\%}$) (higher is better; plotted in logarithmic scale), and test set BLEU (higher is better) on DE$\leftrightarrow$EN of 1M. This is a continuation of Figure \ref{fig:mu-d-freq-bleu}.}
\label{fig:mu-d-freq-bleu-test-x}
\end{figure}

Figure \ref{fig:mu-d-freq-bleu-test-x} contains visualization of $\mu$, $D$, $F_{95\%}$ for BLEU on test set for EN$\rightarrow$DE 1M and DE$\rightarrow$EN 1M.

\begin{figure}[ht]
\begin{subfigure}{\linewidth}
  \centering
  \includegraphics[width=\linewidth,clip]{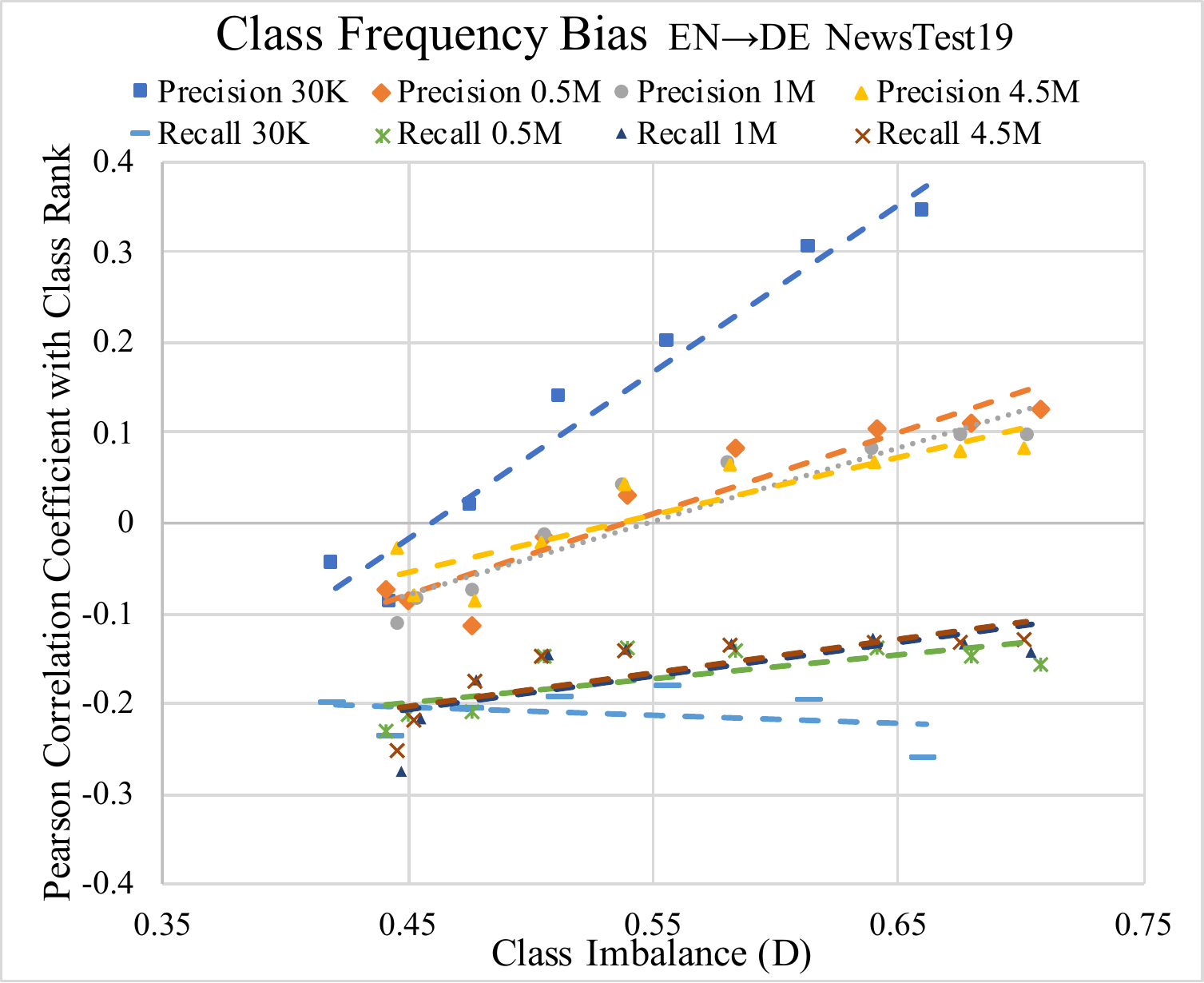}
\end{subfigure}

\begin{subfigure}{\linewidth}
  \centering
  \includegraphics[width=0.99\linewidth,clip]{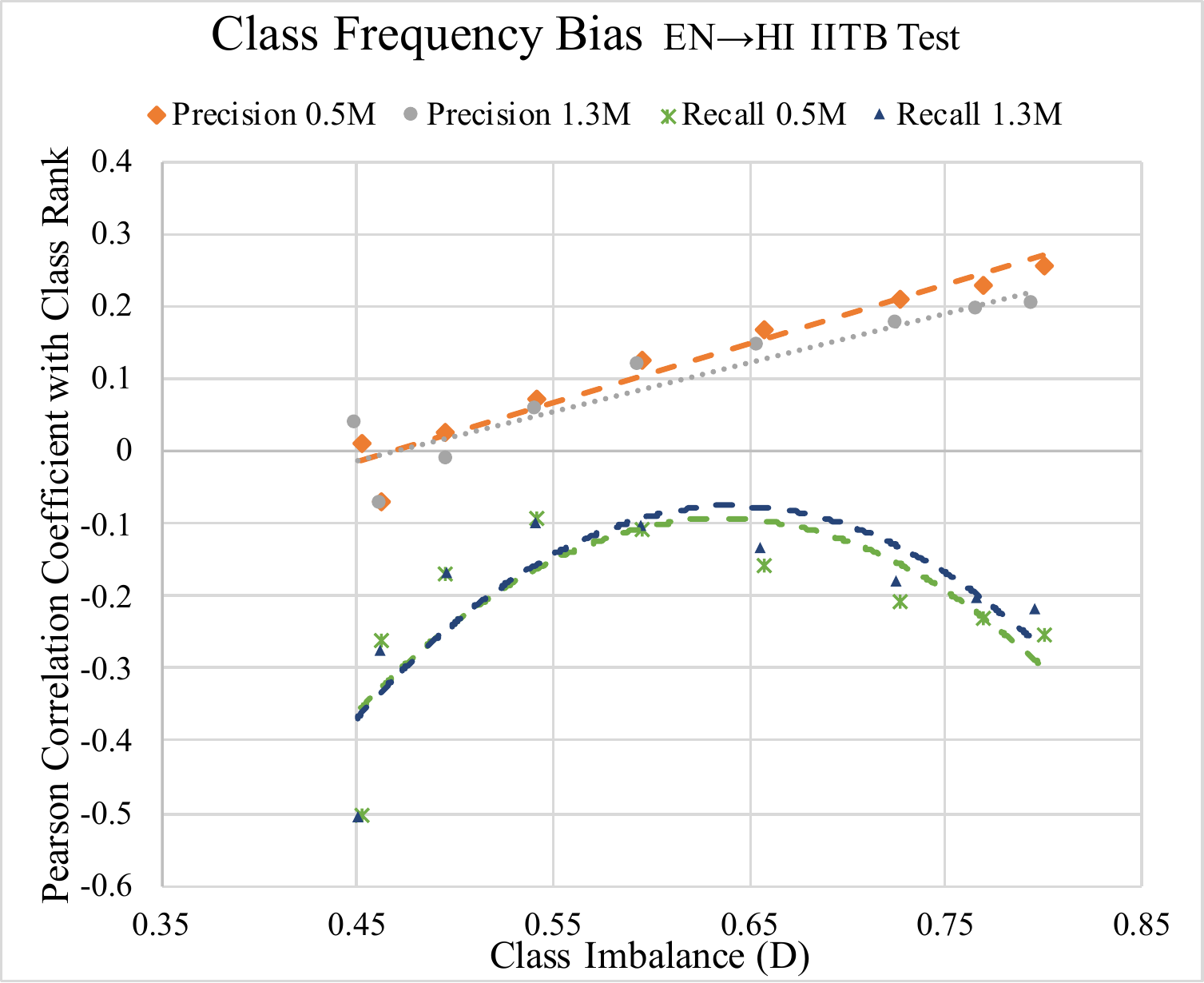}
\end{subfigure}
\caption{Correlation analysis on EN$\rightarrow$DE, and EN$\rightarrow$HI test sets. The non-zero correlation of precision and recall with class rank indicate the class frequency bias in NMT models.}
\label{fig:corr-ende-enhi-test}
\end{figure}

Figure \ref{fig:corr-ende-enhi-test} contains visualization of frequency bias on EN$\rightarrow$DE and EN$\rightarrow$HI languages test sets.

\begin{figure*}[ht]
\begin{subfigure}{\textwidth}
  \centering
  \includegraphics[width=0.6\linewidth,trim={1.4cm 0 0.2cm 16.45cm},clip]{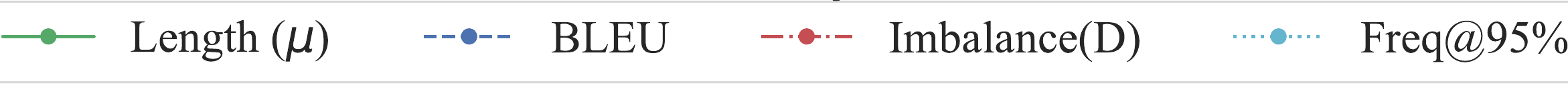}
\end{subfigure}

\begin{subfigure}{.33\textwidth}
  \centering
  \includegraphics[width=0.99\linewidth,trim={2.4cm 1.32cm 4.1cm 0},clip]{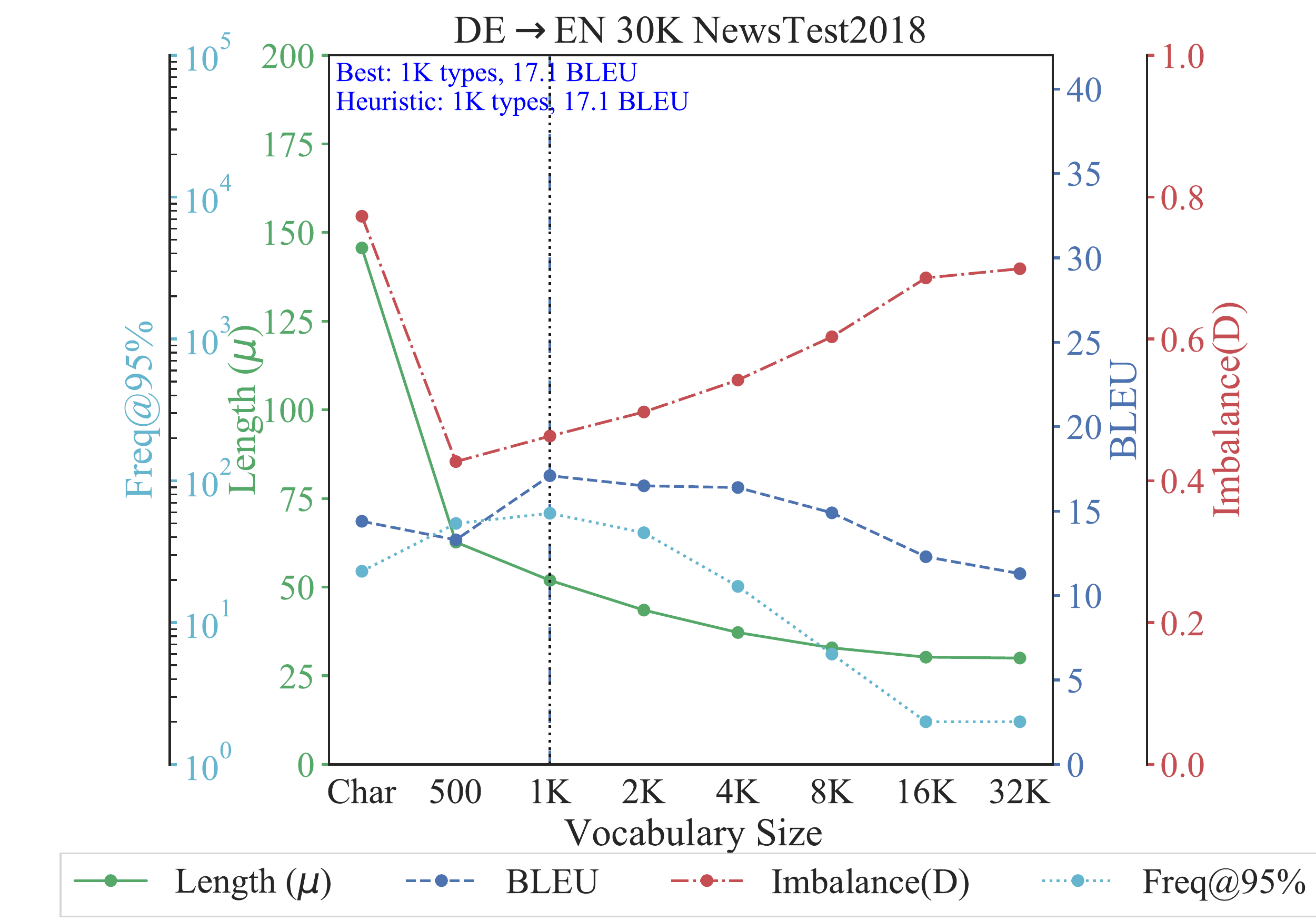}
\end{subfigure}
\begin{subfigure}{.32\textwidth}
  \centering
  \includegraphics[width=0.89\linewidth,trim={5.1cm 1.32cm 4.1cm 0},clip]{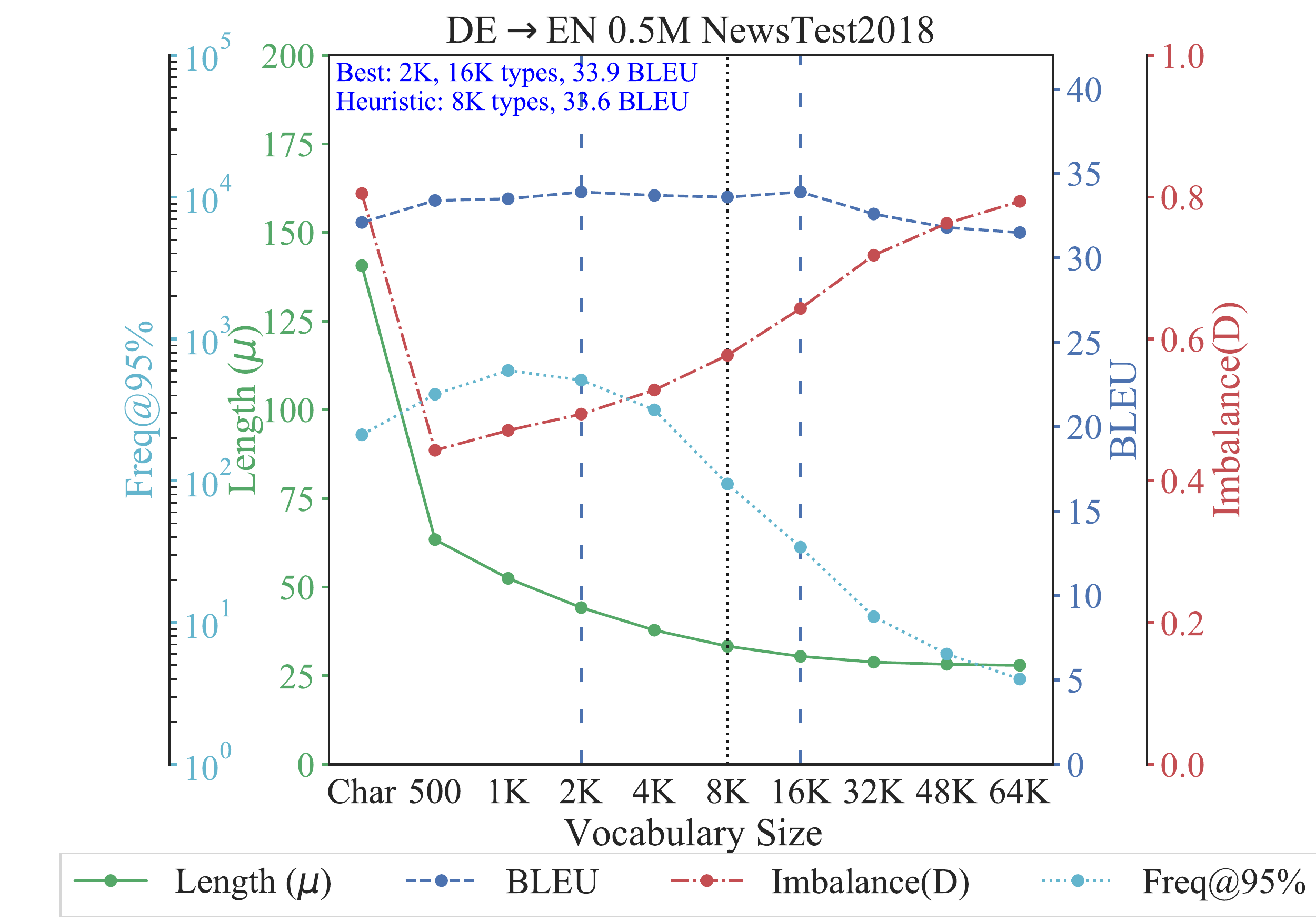}
\end{subfigure}
\begin{subfigure}{.33\textwidth}
  \centering
  \includegraphics[width=0.99\linewidth,trim={5.1cm 1.32cm 1.4cm 0},clip]{4axv-dev-deen-4.5m.pdf}
\end{subfigure}

\begin{subfigure}{.33\textwidth}
  \centering
  \includegraphics[width=0.99\linewidth,trim={2.4cm 1.32cm 4.1cm 0},clip]{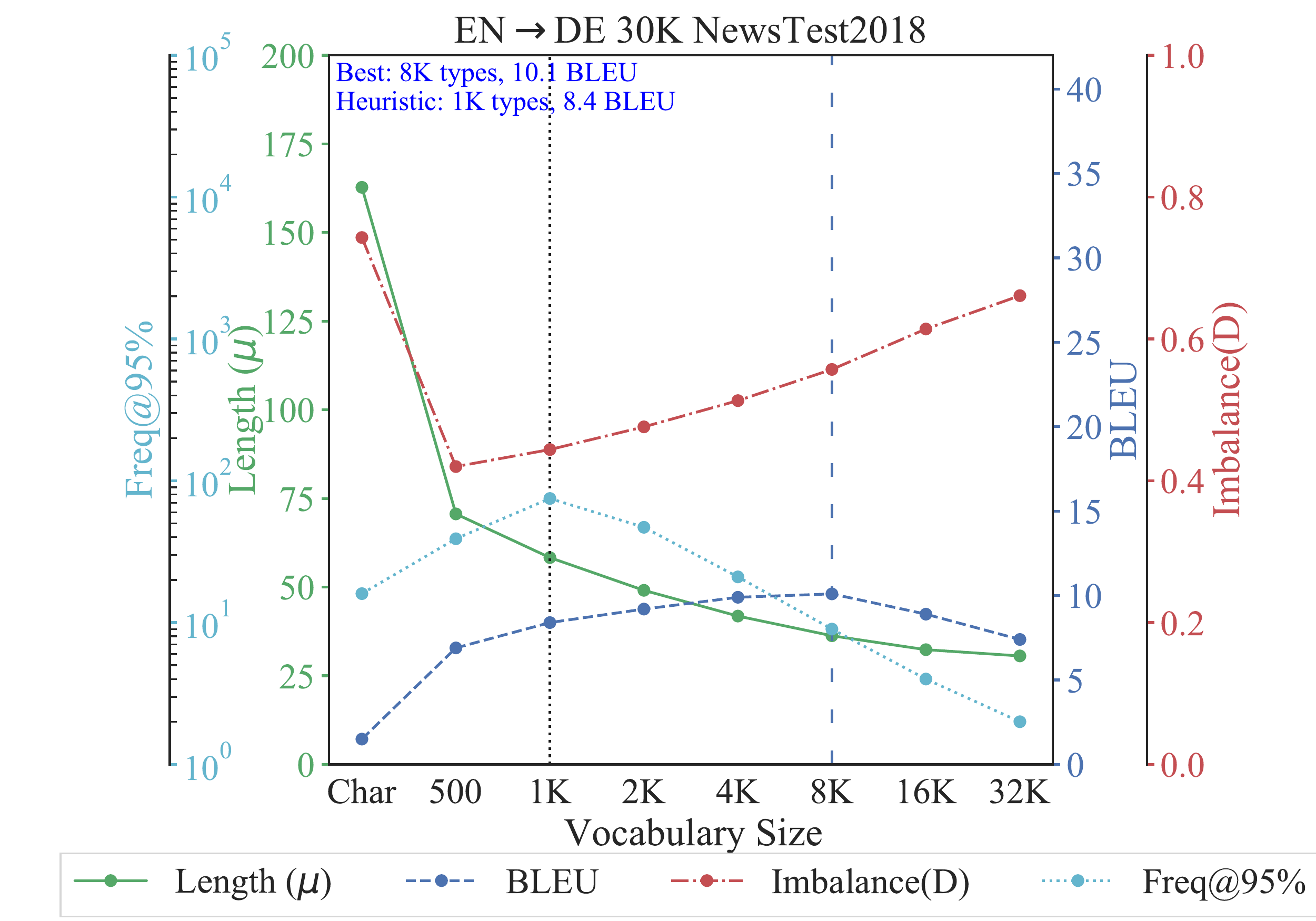}

\end{subfigure}
\begin{subfigure}{.32\textwidth}
  \centering
  \includegraphics[width=0.89\linewidth,trim={5.1cm 1.32cm 4.1cm 0},clip]{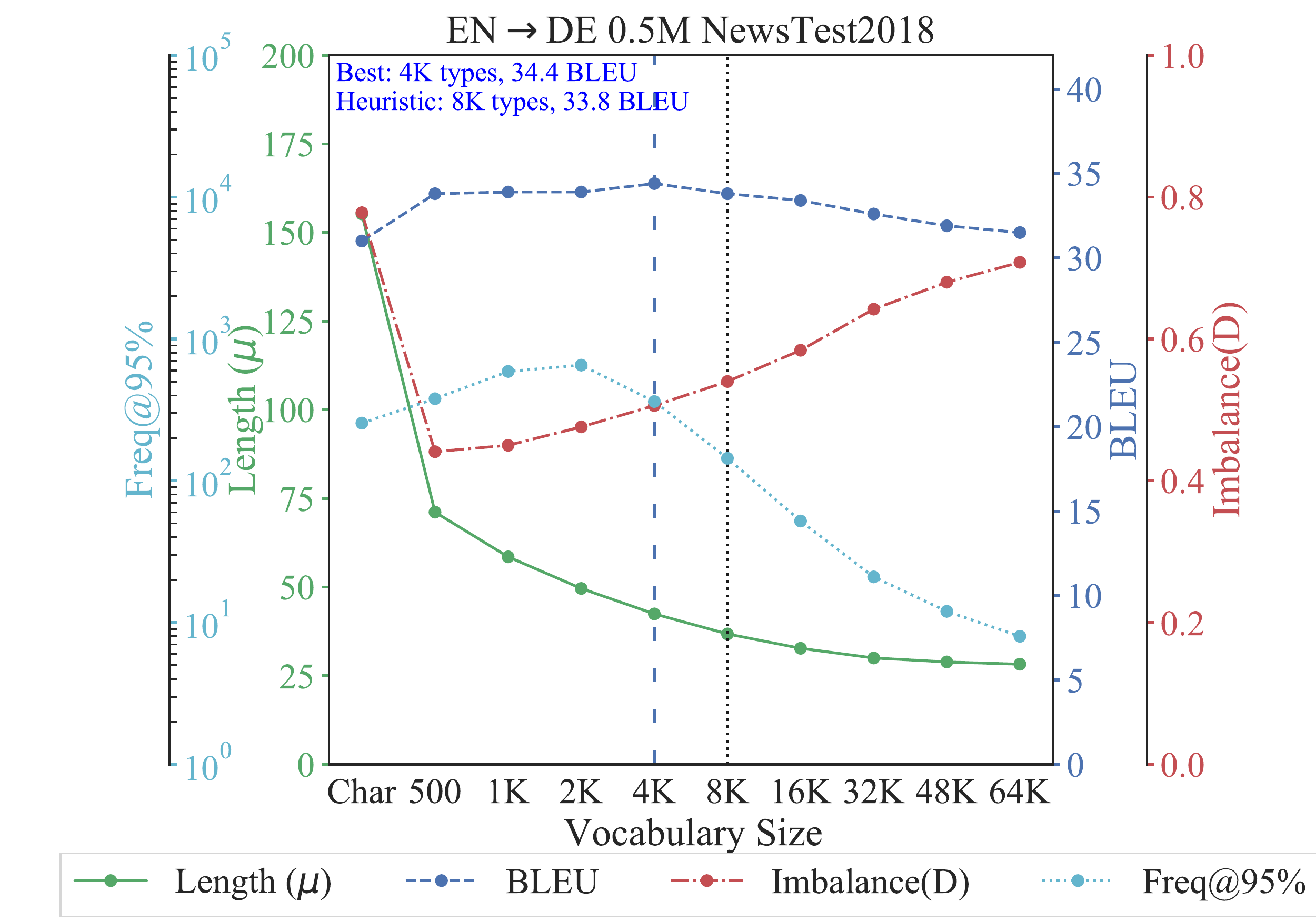}
\end{subfigure}
\begin{subfigure}{.33\textwidth}
  \centering
  \includegraphics[width=0.99\linewidth,trim={5.1cm 1.32cm 1.4cm 0},clip]{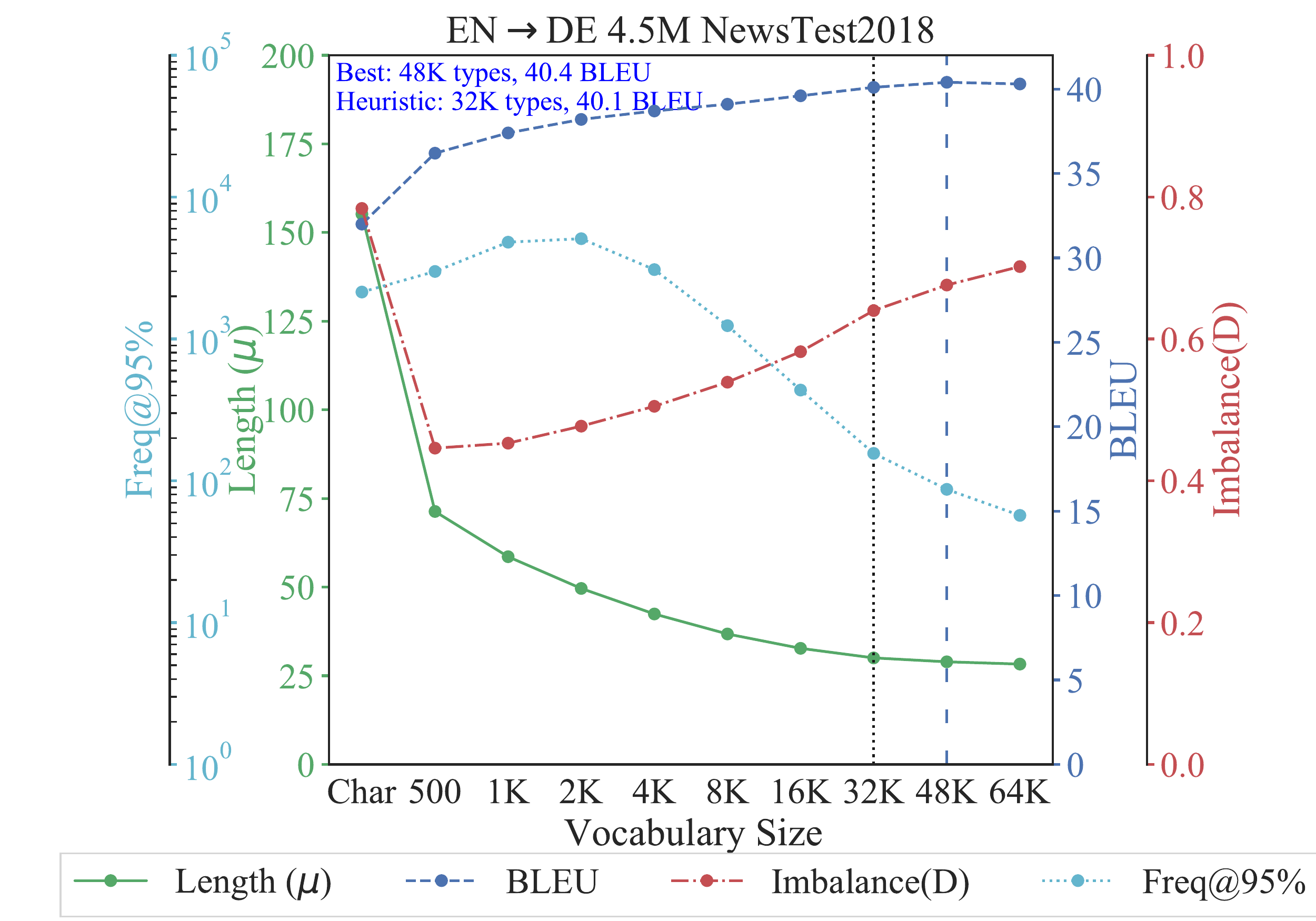}
\end{subfigure}

\begin{subfigure}{.33\textwidth}
  \centering
  \includegraphics[width=0.99\linewidth,trim={2.4cm 1.32cm 4.1cm 0},clip]{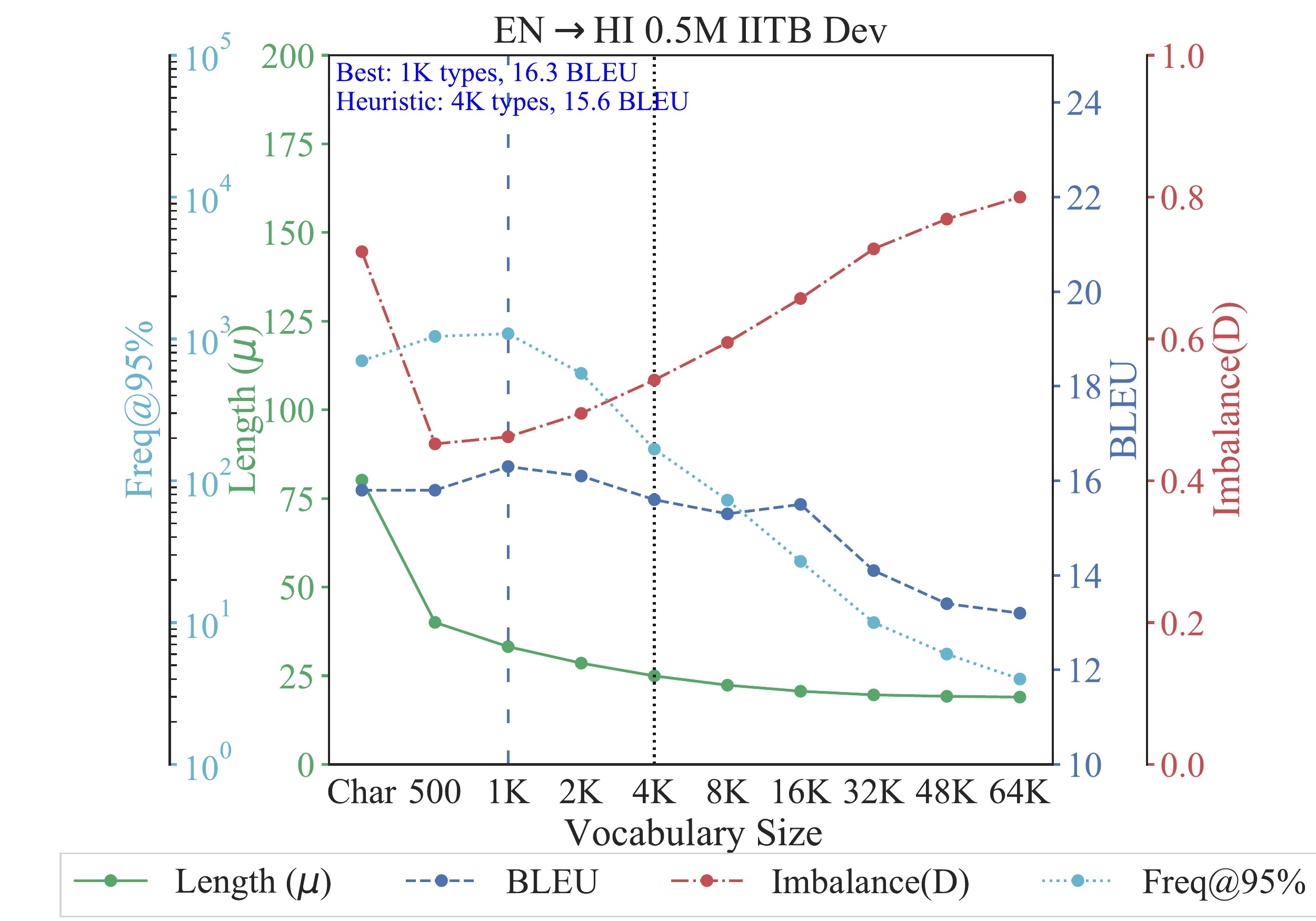}
\end{subfigure}
\begin{subfigure}{.32\textwidth}
  \centering
  \includegraphics[width=0.89\linewidth,trim={5.1cm 1.32cm 4.1cm 0},clip]{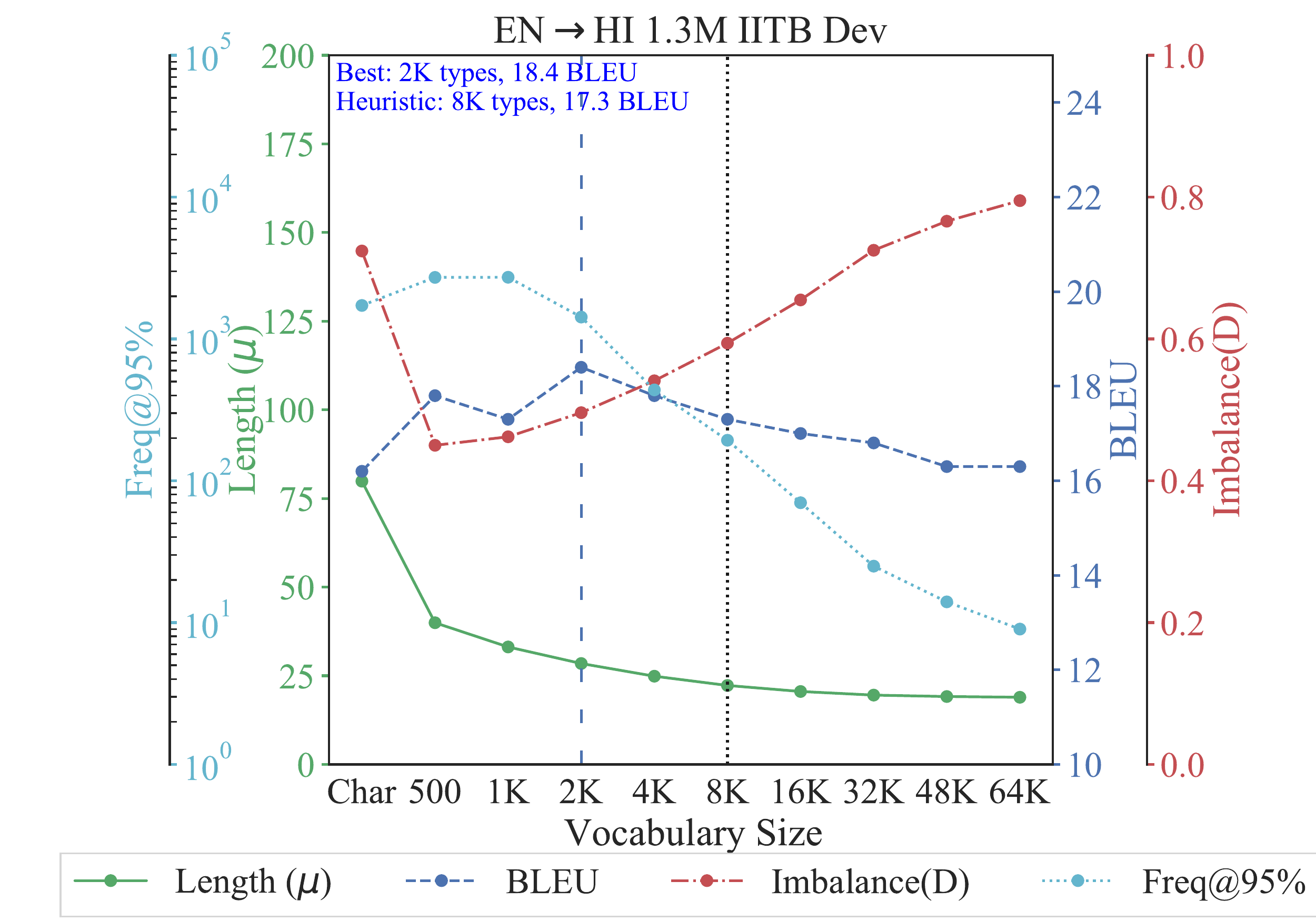}
\end{subfigure}
\begin{subfigure}{.33\textwidth}
  \centering
  \includegraphics[width=0.99\linewidth,trim={5.1cm 1.32cm 1.4cm 0},clip]{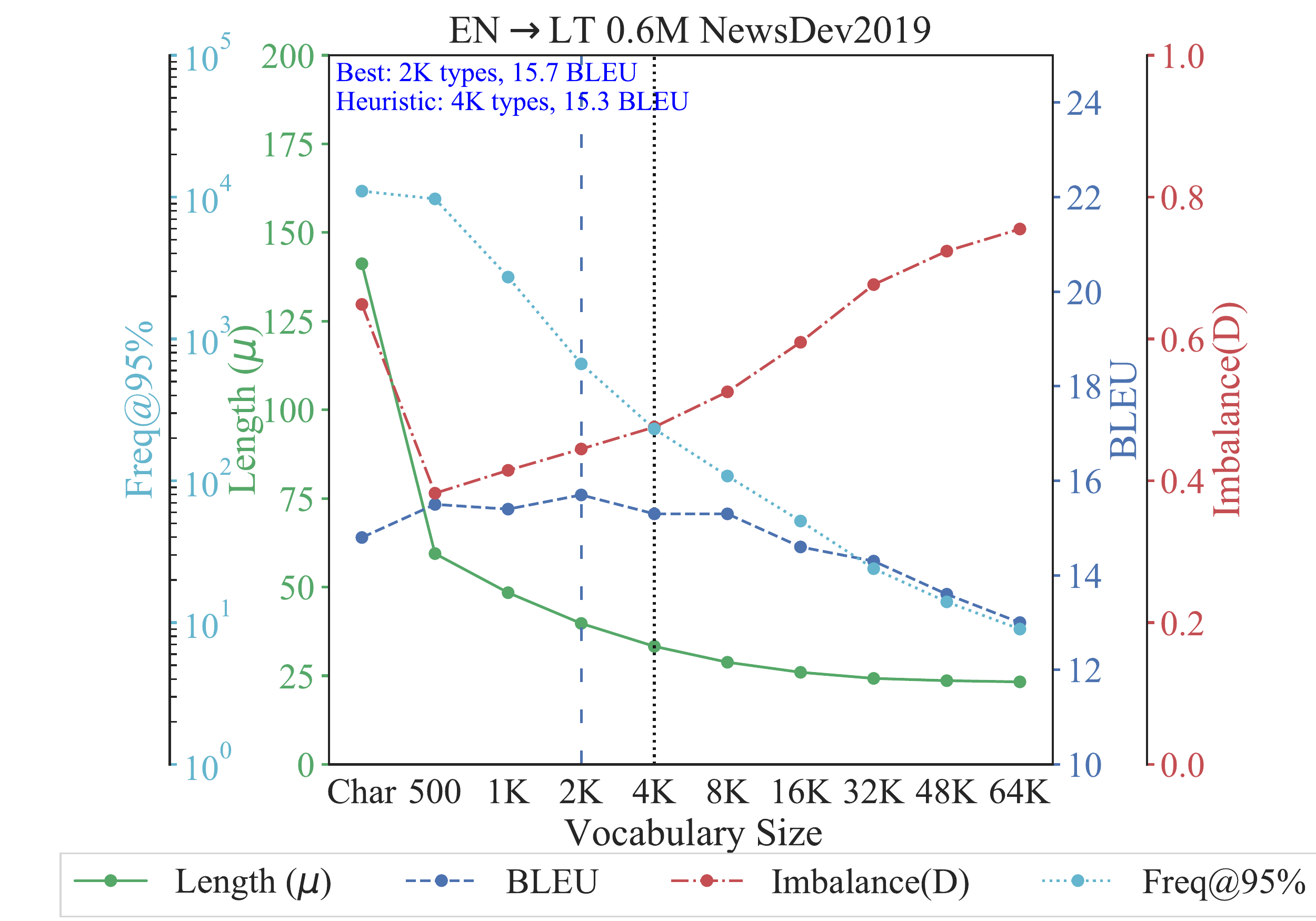}
\end{subfigure}

\begin{subfigure}{.33\textwidth}
  \centering
  \includegraphics[width=0.99\linewidth,trim={2.4cm 1.32cm 4.1cm 0},clip]{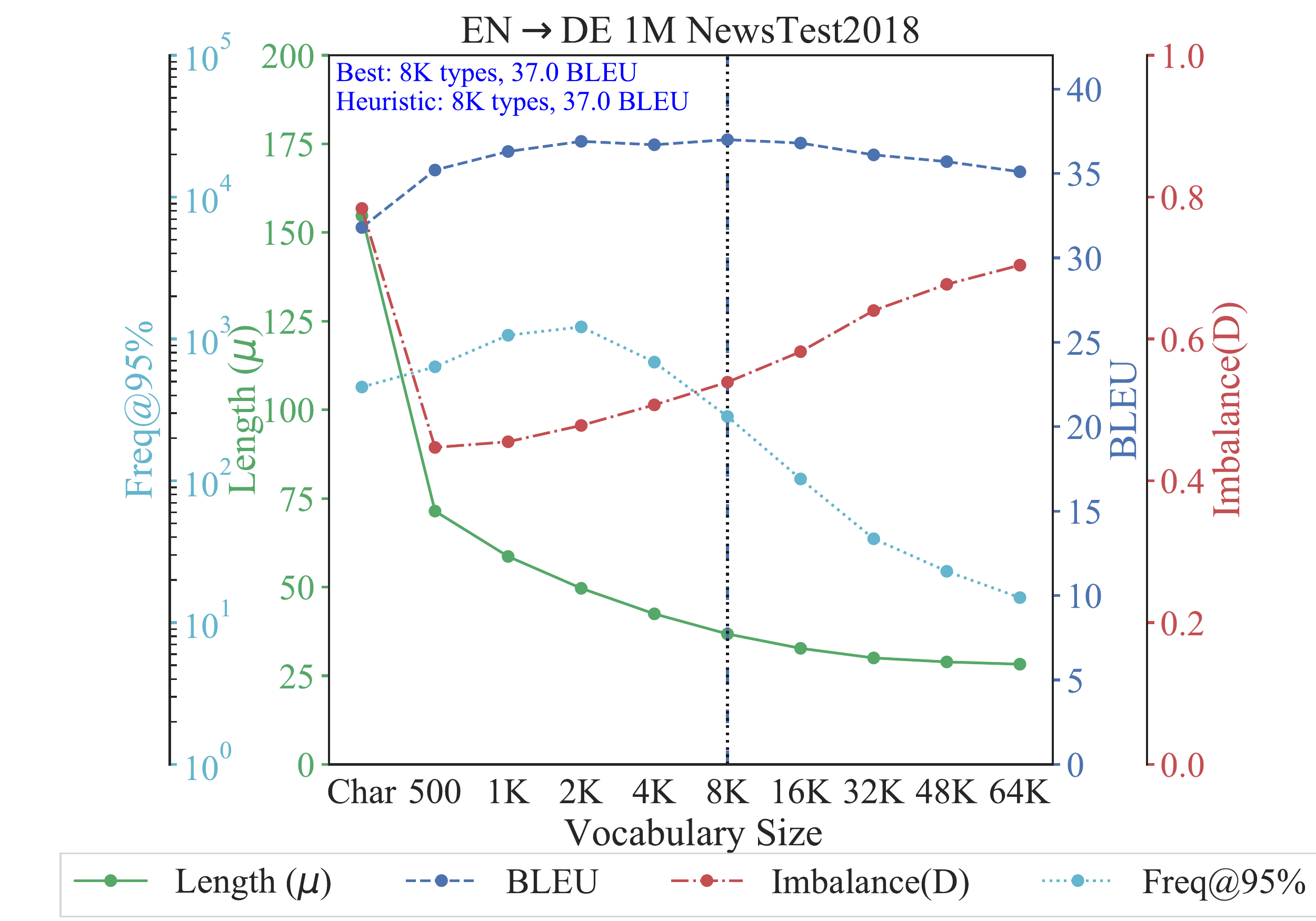}
\end{subfigure}
\begin{subfigure}{.33\textwidth}
  \centering
  \includegraphics[width=0.99\linewidth,trim={5.1cm 1.32cm 1.4cm 0},clip]{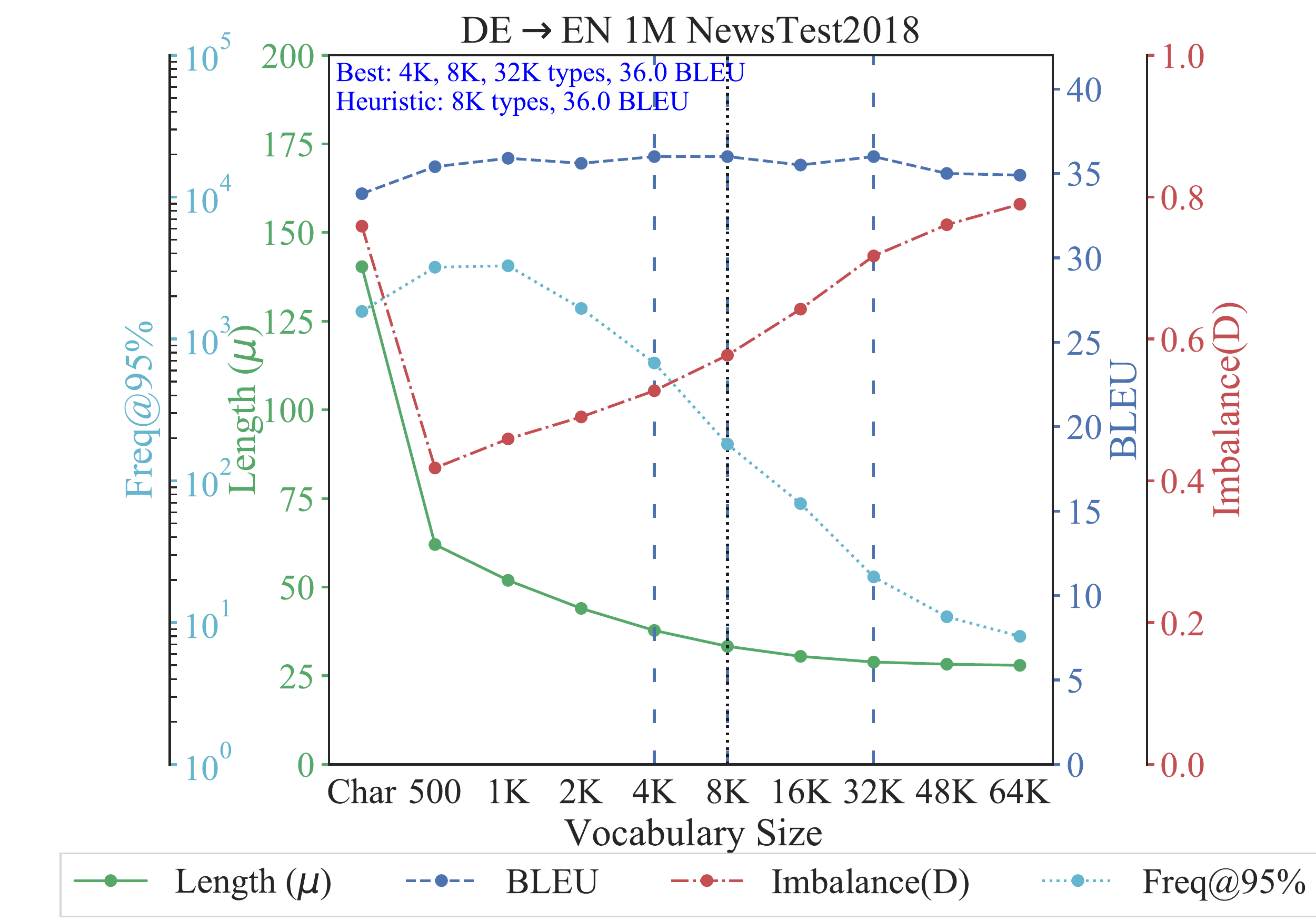}
\end{subfigure}

\caption{Visualization of sequence length ($\mu$) (lower is better), class imbalance (D) (lower is better), frequency of $95^{th}$ percentile class ($F_{95\%}$) (higher is better; plotted in logarithmic scale), and \textit{validation set} BLEU (higher is better) on all language pairs and training data sizes.
The vocabulary sizes that achieved highest BLEU are indicated with dashed vertical lines.}
\label{fig:mu-d-freq-bleu-dev}
\end{figure*}

Figure \ref{fig:mu-d-freq-bleu-dev} contains visualization of $\mu$, $D$, $F_{95\%}$ for BLEU on validation sets.

\end{document}